\documentclass{article}

\usepackage{arxiv}

\usepackage[utf8]{inputenc} 
\usepackage[T1]{fontenc}    
\usepackage{hyperref}       
\usepackage{url}            
\usepackage{booktabs}       
\usepackage{amsfonts}       
\usepackage{nicefrac}       
\usepackage{microtype}      
\usepackage{lipsum}
\usepackage{graphicx}
\usepackage{xcolor}
\usepackage{caption}
\usepackage{multirow}
\usepackage{amsmath}
\usepackage{amssymb}
\usepackage[export]{adjustbox}
\graphicspath{ {./images/} }

\hypersetup{
pdftitle={Synthetic Document Generator for Annotation-free Layout Recognition},
pdfauthor={Natraj Raman}
}
\title{Synthetic Document Generator for Annotation-free Layout Recognition}

\author{
  Natraj Raman$^{1}$, Sameena Shah$^{2}$ and Manuela Veloso$^{2}$ \\
  JPMorgan AI Research\\
  $^{1}$London, UK. \\
  $^{2}$New York, USA. \\
  \texttt{first.last@jpmorgan.com} \\
}

\begin{document}
\maketitle
\begin{abstract}
Analyzing the layout of a document to identify headers, sections, tables, figures etc. is critical to understanding its content.
Deep learning based approaches for detecting the layout structure of document images have been promising.
However, these methods require a large number of annotated examples during training, which are both expensive and time consuming to obtain.
We describe here a synthetic document generator that automatically produces realistic documents with labels for spatial positions, extents and categories of the layout elements. 
The proposed generative process treats every physical component of a document as a random variable and models their intrinsic dependencies using a Bayesian Network graph.  
Our hierarchical formulation using stochastic templates allow parameter sharing between documents for retaining broad themes and yet the distributional characteristics produces visually unique samples, thereby capturing complex and diverse layouts.
We empirically illustrate that a deep layout detection model trained purely on the synthetic documents can match the performance of a model that uses real documents.
\end{abstract}


\section{Introduction}
Documents remain the most popular source of information and machine understanding of the document contents is critical for several tasks such as semantic search, question answering, knowledge base population and summarization. 
Traditional information retrieval techniques~\cite{yang2019simple} from documents tend to focus on the text content and ignore the layout structure. However, the layout cues in documents provide necessary context to the text and can serve as an important signal for measuring information saliency. For instance, the retrieved results for a query phrase that matches both a heading and a typical passage should be ranked in favour of the former, since the headings are a natural concise synopsis of the content.  Many documents also include tabular formats to summarize and compare information. In such cases, detecting the tables, spatially localizing the table cells and capturing the relationship between them is essential for information discovery. Thus identifying the layout elements such as title, section, illustration, table etc. is a necessary step towards effective understanding of a document.

Layout recognition~\cite{binmakhashen2019document} is a challenging problem owing to the variability in the structure and format of documents. Real-world documents originate in multiple domains, follow different templates, have complex structure and may be of poor-quality. In addition, the layout elements are not explicitly encoded in the digital representation of document formats such as PDFs and scanned images. Even if the layout structure is preserved in some document formats, it requires customized parsers and normalization for consistent treatment across the different formats. 

Computer Vision based approaches for layout recognition have the advantage of making no assumption about the document format. Furthermore, the full spectrum of stylistic cues in the visual representation of a document can be efficiently exploited in the image space. Hence it would be beneficial to reframe the task of recognizing the layout elements in a document as detecting the objects in an image. The defacto approach of late in object detection is to use deep learning based feature representation and object segmentation~\cite{liu2020deep}. The success of these methods rely on the availability of a large number of images annotated with the bounding boxes and categories of object instances during training. However, annotation exercises are expensive and even when labeled data is readily accessible, it may cater towards a specific domain or have restricted usage. There is a compelling opportunity to address this desperate shortage of free, diverse, unbiased and open-ended labeled documents by creating them on demand. 

In this work, we address the problem of obtaining labeled documents for layout recognition by automatically generating documents along with ground-truths for spatial positions, extents and categories of the layout elements. Unlike existing  methods~\cite{journet2017doccreator} that require at least a few seed documents, the proposed generation mechanism needs virtually no access to any real documents. Specifically, we treat every component of a document such as a character glyph, geometric shape, font, alignment, color, spacing etc. as a random variable with a prior distribution. The intrinsic dependencies between the variables are modeled using conditional probability distributions. A synthetic document is constructed by sampling the variables from their corresponding distributions and rendering them in an image lattice. In order to account for the commonality in layout structure between subsets of documents, we introduce the notion of stochastic templates, where each template defines a unique set of distributional parameters. Documents that are instances of the same template share parameters and thus retain a broadly similar theme while differing in visual appearance. 

\begin{figure*}[tbp]
\centering
\includegraphics[width=\linewidth]{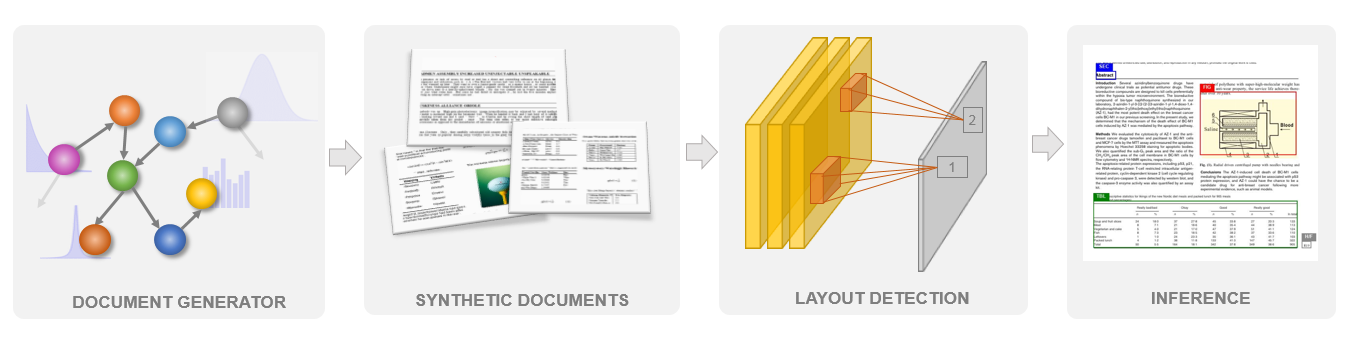}
\caption{Overview of our approach. A document generator encoded as a Bayesian Network is used to construct a large number of synthetic documents. An object detection model is trained purely on these synthetic documents and inference is performed on real documents to identify layout elements such as sections, tables, figures etc. } 
\label{fig_overview}
\end{figure*}

This Bayesian Network~\cite{darwiche2009modeling} formulation of the document generation process can capture complex variations in layout structure, customize the visuals, model uncertainty, and produce large quantities of coherent and realistic synthetic documents. Furthermore, the hierarchical nature of the priors facilitates inductive biases that favor commonly observed patterns and allows parameter inference if samples of these variables from real documents are available. Since the generation process is agnostic to the text content,    it is inherently multi-lingual and domain independent. It also provides flexibility in introducing imaging defects due to printing, optics, scanning, occlusion and degradation for simulation of low quality documents.

We illustrate the utility of our approach (see Figure ~\ref{fig_overview}) by training an object detector network purely on the synthetically generated documents and assess its ability to detect the layout elements such as sections, figures, tables and table cells on real world documents. In particular, we evaluate on three public datasets~\cite{zhong2019publaynet,li2020docbank,zhong2020image} and show that the performance difference between a layout detection model trained on real data and synthetic data is less than 4\%. We also provide intuitive explanations on what the layout recognition model has really learned. Our main contributions are as following:
\begin{enumerate}
    \item We propose a new document generator based on Bayesian Network framework to produce synthetic documents along with ground truths required for layout recognition.

    \item We show that the generator can model complex layouts, customized templates, versatile languages, imaging defects and construct plausible realistic documents.

    \item We empirically illustrate that layout detectors trained simply on synthetic documents can perform as well as those that use real documents.

    \item We include quantitative comparisons, qualitative analysis, ablation studies and visual explanations to substantiate our findings.

\end{enumerate}

Samples of synthetic documents corresponding to different domains are available at \url{https://www.jpmorgan.com/synthetic-data}.

The paper is organized as follows. Section \ref{sec_relwork} compares our work with related efforts, Section \ref{sec_model_synthgen} describes the synthetic generation model in detail, Section \ref{sec_model_objdetect} outlines the layout detection architecture, Section \ref{sec_results} presents samples of generated documents and discusses the recognition results, and finally Section \ref{sec_conclusion} summarizes our findings.

\section{Related Work}\label{sec_relwork}
Layout analysis of documents is a long standing problem of interest with decades of research efforts. Although Optical Character Recognition (OCR) systems~\cite{mori1999optical} meet the requirements of extracting the text content from document images, they do not address the detection and labeling of semantic roles played by the text blocks. Early attempts~\cite{cattoni1998geometric} towards understanding the logical structure of the documents were largely rule-based. They generally segment the documents recursively into distinct geometric regions of homogeneous blocks by applying morphological operations and texture analysis. The main drawback of these methods are that the features are heuristical, require extensive human efforts to tune and fail to generalize. 

Statistical machine learning approaches to model features and train parameters for layout analysis can overcome these drawbacks. Classical algorithms such as Support Vector Machines (SVMs) and Gaussian Mixture Models (GMMs) have been applied~\cite{wei2013evaluation} for document segmentation tasks. However, these methods rely on hand crafted features  and lack the ability to absorb the huge variations of documents in the wild. The advent of deep learning era has paved way for semantically rich features to be automatically extracted from images ~\cite{simonyan2014very, he2016deep}. Deep convolutional network based models have been used to perform layout analysis in ~\cite{gao2017deep}, ~\cite{li2018deeplayout} and ~\cite{li2021few}.  In particular, layout segmentation can be viewed as being equivalent to object detection and the recent successful models~\cite{ren2015faster, lin2017focal} for detecting salient objects from natural scenes can be adapted to identify the layout elements in a document. Following this route, ~\cite{yi2017cnn} focuses on identifying figures and formulae  while ~\cite{prasad2020cascadetabnet} performs table structure recognition. In some cases ~\cite{xu2020layoutlm}, multi-modal information using both the image and text features have also been used for layout recognition. 

The above data-driven deep learning based layout recognition techniques require a large number of annotated images for training, which are both expensive and time consuming to obtain. While it is true that there are a few datasets of documents~\cite{ zhong2019publaynet, li2020docbank,li2020tablebank, zheng2021global} available with ground-truths for the layout bounding boxes, they focus on specific corpora such as scientific publications and are difficult to extend to other domains or customize for new element types. Furthermore, they maybe bound by usage restrictions due to copyright and licensing constraints. In contrast, the proposed method here automatically generates annotated documents for training the deep learning models and hence do not suffer from these limitations. 

Generation of synthetic documents to augment training data has been explored before. In ~\cite{etter2019synthetic}, documents are generated by altering the style attributes in a HTML web-page for training an OCR system that recognized Chinese and Russian characters. Similarly, ~\cite{gupte2021lights} constructs a document synthesis pipeline based on web browser templates to address training data scarcity and mitigate the impact of OCR errors for named entity recognition tasks. ~\cite{journet2017doccreator} proposes a semi-synthetic document creator that alters the font and background of an existing set of real documents to produce new documents. ~\cite{yang2017learning} randomly arranges layout elements in \LaTeX source files and also replaces elements from existing files in an arbitrary fashion to generate synthetic documents.  In contrast to the adhoc nature of above solutions which can only produce limited variations, our approach introduces a principled mechanism to model the physical and logical structure of the document that can create diverse layouts and offers granular control over the depiction of the layout elements.

Image creation using generative neural models has gained interest of late in the vision community.  A recursive autoencoder that maps structural representations based on a limited number of labeled documents is used in ~\cite{patil2020read} to augment training data. From annotated examples of layout bounding boxes,  ~\cite{arroyo2021variational} generates new layouts using a variational autoencoder framework that incorporates attention mechanisms to capture relationships between layout elements. ~\cite{gupta2021layouttransformer} also employ attention networks, but use self-supervised learning similar to masked language models for predicting the ground-truth tokens and modeling the probability distribution of layouts to sample from. The image generator in ~\cite{li2019layoutgan} takes randomly placed graphic elements as input and uses Generative Adverserial Networks (GAN) to produce different layouts with a differential wireframe rendering layer. Recently ~\cite{biswas2021docsynth} proposed an adversarial training approach for controlled document synthesis using reference layout images for guidance. Our work differs from the above methods in its objective, the choice of generative model, its disentangled interpretable nature, the ability to create elements from scratch and importantly, does not require any annotations or real images for generating new documents.

\section{Synthetic Document Generator}\label{sec_model_synthgen}
This section describes the Bayesian Network generative model from which the synthetic documents are sampled.

\subsection{Bayesian Network Formulation}
A document is comprised of primitive graphical units such as character glyphs, pictures and geometric shapes that are assembled in a logical structure. These primitive units are governed by a set of style attributes that define their visual representation. For example, the font, alignment and spacing attributes influence the appearance of character blocks while stroke attributes control the thickness and color of a line or rectangular shape. These units are grouped into logical layout elements such as headers, sections, tables, footers etc.  that describe the semantics of the graphical elements. 

The diversity in observed documents is as a result of variations in the primitive units, their styles and the positional structure of the layout elements. A good document generator should have the capability to model these complex variations, provide flexibility to tailor the visuals, account for noise and produce realistic documents.  

Our proposed solution employs a Bayesian Network to define the synthetic document generation process. The primitive units, style attributes and layout elements are all treated as random variables and represented as nodes in a graph. The intrinsic dependencies between these variables are symbolised using directed edges. Every node variable is endowed with a conditional probability distribution that quantifies the strength of relationship between a node and its parents. Parent-less nodes are described by marginal distributions. The parameters of a distribution may themselves have prior distributions with hyper-parameters resulting in a hierarchical model. These hyper-parameters are automatically learned during posterior inference if a corpus of real documents is available or set to pre-defined values if expert knowledge is available or could even have further vague priors, reflecting lack of information.

We define stochastic templates that vary in the value of distributional parameters in
order to capture the commonality in layout structure across different domains. For instance, a scientific document may have a visual appearance that is totally distinct from a corporate report, yet the documents within the scientific domain may have common patterns. The templates help encode the intra-domain similarity by sharing the same hyper-parameter values for a variable, while the inter-domain similarity is simulated by using unique values across the templates.  

The variables that govern the composition of a document are sampled from this Bayesian Network and rendered as an image to produce a synthetic document. The bounding boxes of the layout elements and their corresponding category are recorded during the generative process, thereby creating an implicitly annotated document.  This formulation of automatic document synthesis using a  Bayesian Network has the following advantages:
\begin{itemize}
    \item The distributional characteristics of the generative process captures the complex variations in layout, style, content, structure and even abnormalities that are typical in the wild.
    \item The definition of stochastic templates offer a mechanism to customize for domain specific visual appearances and extend to new unseen domains.
    \item The hierarchical nature of the priors provides flexibility in learning from data,  bias towards a-priori beliefs and focus on discriminative features.   
    \item The causal well-defined structure of the network can cheaply produce massive quantities of coherent and realistic documents, without the need for expensive infrastructure.
     
\end{itemize}

\begin{figure*}[!tbp]
\centering
\includegraphics[width=\linewidth]{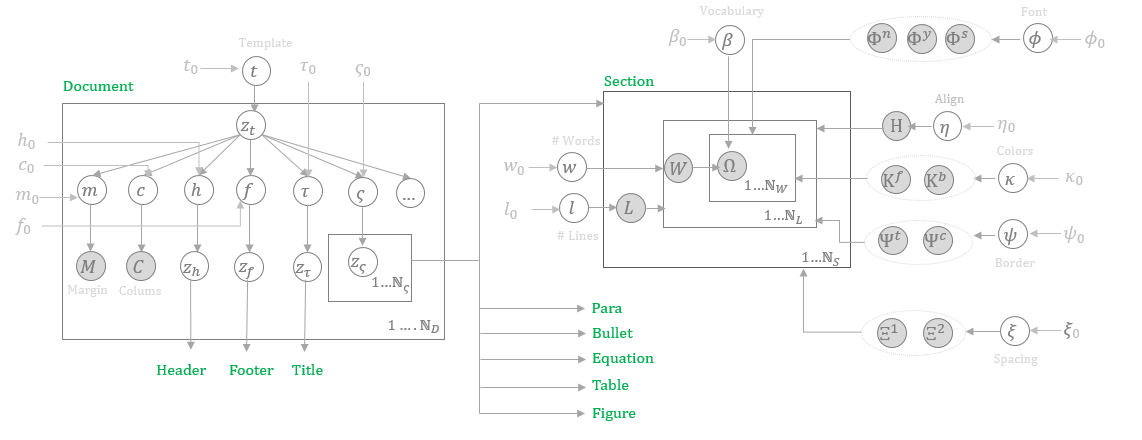}
\caption{Cross section of the Bayesian network in plate notation. The shaded variables are observed and priors are denoted with subscript zero. The subnetworks and functions of variables are marked for illustration. } 
\label{fig_bayesnet1}
\end{figure*}

\subsection{Generative Process}
A Bayesian Network $\mathcal{G}=(\mathcal{V},\mathcal{E})$ is a directed acyclic graph with $\mathcal{V} \in \{v_i\}_{i=1}^N$ being the set of random variables represented as graph nodes and $\mathcal{E}$ being the set of edges that define the parent-child relationship between any two nodes. The random variables may be discrete or continuous, observed or latent and potentially multivariate. We use small letters such as $x$ to represent a latent random variable and capital letters such as $X$ to denote an observed variable. Indicators are a special type of latent variable that provides an index over the possible values of a discrete variable and are referred as $z_x$. The parameters that govern the prior distribution of a variable uses subscript zero. Thus $x_0$ is the set of (hyper) parameters for $x$. We embed these parameters within the classical notation of a distribution - for example, if $x$ is distributed as a Gaussian, then we write $\mathcal{N}(\mu_{x_0}, \sigma_{x_0}^2)$. Finally, $\mathbb{N}_x$ denotes the number of times a variable $x$ is repeatedly sampled while $N_x$ is the number of possible values of a discrete variable $x$. 

For exposition convenience, we decompose the network into a set of subnetworks $\mathcal{G}=\{g_1 \cup...\cup g_j \cup ... \cup g_J\}$ where each $g_j$ defines a network fragment. Typically, there is a separate subnetwork for each of the layout element. While the variables are not shared between the element subnetworks, it may share variables that are common to a document. Figure \ref{fig_bayesnet1} provides a cross-section of the network with emphasis on the document and section subnetworks.

\subsubsection{Document Subnetwork}
The generative process for a document $d$ proceeds as follows:
\begin{enumerate}
    \item \emph{Sample template}: Let $t \in \mathbb{R}^{N_t}$ be a set of multinomial probabilities over $N_t$ different templates and be Dirichlet distributed with parameters $\alpha_{t_0} \in \mathbb{R}^+$. Choose a template $z_{t,d}$ as
    \begin{align}
    \begin{split}
    t_d &\sim Dir(\alpha_{t_0})
    \\
    z_{t,d} &\sim Mult(t_d).
    \end{split}    
    \end{align}
   Each template defines a unique set of parameters for a given prior distribution i.e. there are $N_t$ values for an $x_0$. In the following, the template index $z_{t,d}$ is used to select the hyper-parameter corresponding to a document as $x_{0,z_{t,d}}$. 
   
   \item \emph{Sample document level variables}: The variables such as margin, number of columns and background color influence the overall visual of a document, and are applicable at a document level. Let $\mu_m$ and $\sigma_m^2$ be the mean and variance of a real valued margin variable $m$. Let $\mu_m$ be bestowed a normal prior with mean $\mu_{m_0}$ and variance $\sigma_{m_0}^2$, and $\sigma_m^2$ with an inverse gamma prior parameterized in terms of shape $a_{m_0}$ and scale $b_{m_0}$. Choose a margin $M_d$ as 
    \begin{align}
    \begin{split}
    \mu_{m,d} &\sim \mathcal{N}(\mu_{m_0,z_{t,d}}, \sigma_{m_0,z_{t,d}}^2)
    \\
    \sigma_{m_d}^2 &\sim InverseGamma(a_{m_0,z_{t,d}}, b_{m_0,z_{t,d}})
    \\
    M_d &\sim \mathcal{N}(\mu_{m,d}, \sigma_{m,d}^2).
    \end{split}    
    \end{align}
    
    Similarly, let $c$ be a discrete variable representing the different choices for number of document columns. The observed column value in a document is sampled as
    \begin{align}
    \begin{split}
    c_d &\sim Dir(\alpha_{c_{0,z_{t,d}}})
    \\
    C_d &\sim Mult(c_d).
    \end{split}    
    \end{align}
    
    The sampling of background follows the same procedure as $C$.
    
    \item \emph{Sample shared variables}: Variables are shared across the subnetworks to provide an opportunity to maintain a consistent look and feel across the document, while retaining the option of customization with subnetwork specific overrides.  These shared variables include font name, font size and text color. The discrete variables font name and text color are sampled similar to (3). The font size variable follows a two parameter exponential distribution that in general discourages large sized fonts. Let $\phi^s$ be a font size variable with known location $\theta_{\phi^s}$ and scale $\lambda_{\phi^s}$, which is Gamma distributed with shape $a_{\phi^s_0}$ and scale $b_{\phi^s_0}$. Choose a positive real-valued font size for the document as
    \begin{align}
    \begin{split}
    \lambda_{\phi^s,d} &\sim Gamma(a_{\phi^s_0,z_{t,d}}, b_{\phi^s_0,z_{t,d}})
    \\
    \Phi^{s}_d &\sim Exp(\theta_{\phi^s}, \lambda_{\phi^{s},d}).
    \end{split}    
    \end{align}
    
    \item \emph{Sample single instance elements}: Let $h$ be a Bernoulli random variable that denotes the probability of a header element with a Beta prior parameterized in terms of shape $a_{h_0}$ and scale $b_{h_0}$.  The presence of a document header is sampled as 
    \begin{align}
    \begin{split}
    h_d &\sim Beta(a_{h_0,z_{t,d}}, b_{h_0,z_{t,d}})
    \\
    z_{h,d} &\sim Ber(h_d).
    \end{split}    
    \end{align}
    Similarly, the presence of other elements such as footer $f$ and title $\tau$ that occur only once in a document are sampled. When an element is present in the document, the subnetworks corresponding to the element are instantiated and sampled.
    
    \item \emph{Sample multi-instance elements}: Let $\zeta$ be a discrete variable that represents the elements such as section, table, figure, paragraph, bullet, equation etc. that can occur multiple times in a document. First sample the probabilities of these elements in a document from its Dirichlet prior and then repeatedly sample an element as follows:
    \begin{align}
    \begin{split}
    \zeta_d &\sim Dir(\alpha_{\zeta_{0,z_{t,d}}})
    \\
    z_{\zeta_{d,n}} &\sim Mult(\zeta_d)  \qquad \forall n=1...N_{\zeta}
    \end{split}    
    \end{align}
    
    where $N_{\zeta}$ is a Poisson distributed variable. The subnetwork corresponding to an element defined by $z_{\zeta_{d,n}}$ is repeatedly instantiated and sampled as described in the sequel.
    
\end{enumerate}
 
\subsubsection{Section Subnetwork}\label{subsec_secsubnet}
The section subnet may be instantiated multiple times, and hence when generating a section, the variables that are shared across all sections in the document are sampled once from their priors first. These include the Dirichlet distributed font styles $\phi^y$, horizontal alignment $\eta$, fore color $\kappa^f$,  back color $\kappa^b$, border type $\psi^t$ and border color $\psi^c$ along with the uniformly distributed continuous variables for font scale $\phi^c$, pre section spacing $\xi^1$ and post section spacing $\xi^2$. The above continuous variables are treated as relative values that are scaled against a base unit - for instance, the spacing after a section is a product of a base document level spacing and the post section spacing. Furthermore, a discrete variable $l$ to denote the number of lines in a section and a real valued Gaussian distributed variable $w$ for modeling the number of words in a line is introduced. Finally, we treat words as fundamental units for character blocks and maintain a pre-specified vocabulary of words with multinomial probabilities $\beta$ and sample these probabilities from a Dirichlet variable $\beta_0$ once per document.  

The content of a section $s$ is generated by first sampling the number of lines $L$ in a section, then the number of words $W$ for each line in the section and finally the word $\Omega$ from the vocabulary.

\begin{align}
\begin{split}
L_{d,s} &\sim Mult(l_d)
\\
W_{d,s,i} &\sim \mathcal{N}(\mu_{w_d}, \sigma_{w_d}^2) \qquad \forall i = 1 ... L_{d,s}
\\
\Omega_{d,s,i,n} &\sim Mult(\beta) \qquad \forall i = 1 ... L_{d,s}, \quad \forall n= 1 ... W_{d,s,i}
\end{split}
\end{align}

This section specific content is rendered by applying the sampled pan-section style variables.

\subsubsection{Table and Figure Subnetworks}
A table is parameterized by its percentage of document width, alignment, borders, horizontal and vertical padding, pre and post spacing across blocks, and the number of rows and columns. We also introduce variables to define font and alignment attributes that are unique to a row, column or even a table cell. The cell dimensions are parameterized by a Dirichlet distributed variable that indicates the percentage of width for a cell and a categorical variable denoting the number of lines. Thus a cell maybe empty, has a single value or is wrapped with multiple lines. The padding variables follow an exponential distribution similar to (4) while we assign a fat-tailed Cauchy prior to the number of table rows to allow significant probability for values away from the mean. Rest of the variables follow a Dirichlet or Gaussian prior depending on whether they are discrete or continuous. A table cell content is sampled similar to equation (7).

The figures include both natural images and automatically generated synthetic images in the form of charts. For the former, we maintain a library of images and perform a uniform sampling over this library while new variables for the number of sub-plots within a chart figure, and the chart type within each sub-plot are introduced for the latter. Example chart types include bar, line, scatter, pie, heatmap etc. The chart data  is generated randomly from a uniform distribution. Finally, the figure dimensions are parameterized with two continuous variables. The prior definitions follow a pattern similar to the other subnetworks.

Both the table and figure subnetworks include a caption subnet that includes variables for the caption position and content length. The text content of the caption is generated as in (7). 

\subsubsection{Other Subnetworks}
The title subnetwork follows the same procedure as listed in Section \ref{subsec_secsubnet} except that there is only a single title per document. The distributional parameters of the variables however vary from that of the section, allowing to introduce title specific nuances. 

The paragraph subnetwork introduces new variables for the number of lines in a para and the spacing between  lines and across blocks. These continuous variables follow a bounded uniform distribution and are sampled each time the network is instantiated. The content in a line is sampled from a pre-defined corpus of sentences, once per line.  The bullet subnetwork is similar to the paragraph, except that a discrete valued bullet type variable and a continuous bullet margin offset variable are used to simulate enumerated text.

The header (and footer) subnetwork contains additional variables for the number of columns in a header and the type of content within a column. For instance, the header may contain a 3 column layout of equal spaces with the first column containing a short logo type text, the second column containing a page number and the third column with a text denoting a running title. The above content types define their corresponding style variables for font, color and alignment and follow the procedure described in Section \ref{subsec_secsubnet}. 

\subsubsection{Defect Modeling}
Documents with imperfections are common in unconstrained environments due to physical degradation and scanning errors. We simulate such low quality documents by introducing defects deliberately into the generative process. Specifically, a series of parameterized deformations corresponding to effects such as ink seepage, occlusion, shadows, blurring and damaged corners are defined and these parameters are treated as random variables as before.  For instance,  a digital watermark effect has variables for the text content, rotation angle and location and these variables are  sampled similar to the other subnetworks.     

\subsubsection{Image Synthesis}\label{subsec_imgsynth}
Finally, given the set of observed values sampled from the network, an image construction library is used to draw the primitive graphical units at appropriate locations by applying the sampled style attributes. Simultaneously, the position of the corresponding layout elements are recorded. This results in a document $d$ defined on an image lattice $I_d$ with layout elements $\mathbf{L_{d}}=\{(\iota_{i}, bbox_{i})\}_{i=1}^{N_{L_d}}$, where $N_{L_d}$ is the number of layout elements in document $d$, $\iota_i \in \{O_1...O_{\mathbf{C}}\}$ is one of the $\mathbf{C}$ layout categories and $bbox_i \in \mathbb{R}_{+}^4$ denotes a bounding box.

\subsection{Parameter Inference}
The values of the observed variables in the network can be sampled directly from the conditionals and their corresponding priors. However, it is efficient to learn these parameters offline if we have a corpus of real-world documents. We consider a partially observed network setting where we have data samples for subsets of variables and exploit the  knowledge about probabilistic influences between the variables in order to perform posterior inference. Let $\mathcal{D} = \{\chi^1....\chi^M\}$ be a multiset of observed data samples for $M$ different variables where $M \ll N$. Let $\Theta_i$ denote the parameters of the distribution of an observed variable $i$ with $\Pi_i$ being the parameters of its parent variable. The conditional independence  of the variables encoded in the above described Bayesian Network structure results in the factorization of their joint distribution into a product of locals and consequently the posterior distribution for $i$ is given as 
\begin{align}
p(\Theta_i|\chi^i) = \int p(\chi^i|\Theta_i) p(\Theta_i|\Pi_i) d\Theta_i.
\end{align}
Our specific choice of the distributions for the conditionals and their corresponding priors ensure that the posterior distribution is in the same form as the prior. Due to these conjugate priors, the posteriors can be computed in a closed form. For instance, a multinomial conditional has a Dirichlet prior with parameters $\alpha_{i_0} \in \mathbb{R}^K$, and from data samples $\chi^i$ we can compute $(N^{i1},...,N^{ik},...N^{iK})$ where $N^{ik}$ is the number of observations for a discrete variable $i$ with value $k$. The posterior parameters can now be  sampled from $Dir(\alpha_{i_*})$ where $\alpha_{i_*,k}=\alpha_{i_0,k}+N^{ik}$. Similar analytical updates for the posteriors are available~\cite{diaconis1979conjugate} for the other conditionals including a Bernoulli variable with Beta prior, an exponential variable with Gamma prior, a Gaussian variable with normal prior for the mean and inverse Gamma prior for the variance.

\section{Document Layout Recognition}\label{sec_model_objdetect}

\begin{figure*}[!tbp]
\centering
\includegraphics[width=\linewidth]{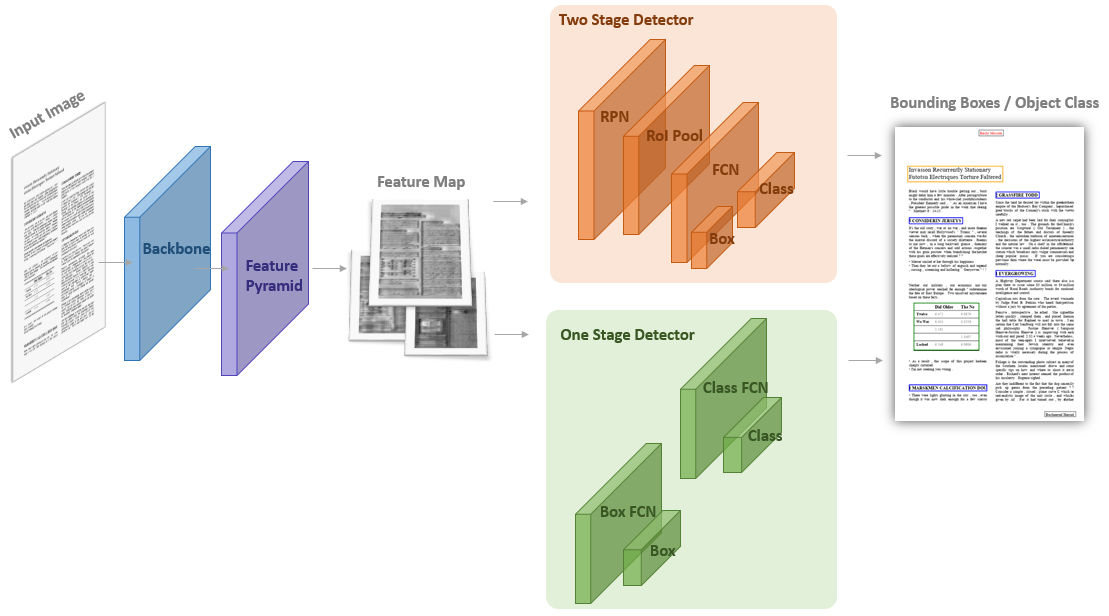}
\caption{Layout Recognition Model Architecture. A feature extraction network takes an image of arbitrary size as input and produces feature maps at different scales. An object detector network determines the categories and bounding boxes of the layout elements.} 
\label{fig_objdet1}
\end{figure*}

The goal of layout recognition is to locate the various layout elements in a rasterized document and identify their corresponding layout categories. This objective can be cast as an object detection problem, where instances of visual objects of a certain class along with their spatial location and extent are retrieved from an image. The various layout elements in the document are characterized as the objects of interest here.  

Deep learning based feature representation and object segmentation have emerged as the defacto approach~\cite{liu2020deep} to address the object detection problem. However, training these models require a large number of images annotated with the bounding boxes and classes of object instances. The document generator described above can cheaply produce synthetic labeled images required for training a deep learning based object detector, thereby avoiding the need for expensive annotation procedures.

The object detection model architecture used for layout recognition is outlined in Figure ~\ref{fig_objdet1}. It consists of a backbone deep ConvNet~\cite{he2016deep} that computes a feature hierarchy for a given image at several scales with multiple layers of convolutions and downsampling. This bottom-up pathway of feed-forward computations captures semantics from low to high level and facilitates scale invariance.  The lower level semantics produced by this feature hierarchy is combined via lateral connections with a top-down pathway that upsamples spatially coarser but semantically stronger features. The resulting feature pyramid network~\cite{lin2017feature} produces a feature map with fixed dimensions that captures high-level semantics at multiple resolution scales.

This rich multi-scale feature map is fed as input into an object detector subnetwork that performs object classification and bounding box regression. We consider two main classes of detector architectures: (i) a two stage detector in which a sparse set of candidate object locations are generated first and subsequently these candidate locations are refined and classified and (ii) a one stage detector which performs classification and regression directly on a regular dense sampling of object locations. The two-stage detector includes a region proposal network (RPN)~\cite{ren2015faster} that applies convolutions to the feature map and produces bounding boxes for object proposals along with objectness scores. In addition, a region of interest (RoI) pooling layer extracts fixed length feature vector from the feature map corresponding to the object proposals. Finally, these feature vectors are fed into a sequence of fully connected networks (FCN) to determine the object classes and their bounding boxes. In contrast, the one-stage detector~\cite{lin2017focal} avoids explicit region proposals and instead considers locations across the image. It uses two parallel deep subnetworks directly on the feature map, one to predict the probability of object presence at a spatial position and another for box regression. 

Given the set of synthetic document images $\mathbf{I}=\{I_1....I_{N_D}\}$ and their corresponding labels $\mathbf{L}$, the layout recognition model is trained based on the above architecture. We compare in our experiments the recognition performance on real documents for different choices of the feature extractor and object detector.

\section{Experiments}\label{sec_results}
We present the output of the generator model, outline the evaluation settings and finally discuss the layout recognition results in this section.

\subsection{Document Generator Results}
A few examples of the synthetic documents produced by the document generator is shown in Figure ~\ref{fig_synthdocs}. The variety, coherence and realistic nature of the generated documents is evident. The documents have different styles corresponding to fonts, colors, borders, alignments and content lengths for the title and section headings. Not all documents contain these layout elements, mimicking mundane descriptive pages present in typical documents. The bottom part of the figure illustrates the wide range of table element representations. The table borders pertain to rows, columns, headers or individual cells for grid type appearance. The table size, number of rows and columns, density in layout and cell alignments all differ across the tables. The documents also include both natural~\cite{deng2009imagenet,afifi2019afif4} and synthetically generated multi-chart images. Furthermore, the documents vary in their number of columns and the presence, location and style of page numbers,  running titles and marker glyphs in the header and footer elements.

\begin{table*}[tbp]
\centering
\begin{tabular}{ccc}
\frame{\includegraphics[width=.3\linewidth,valign=t]{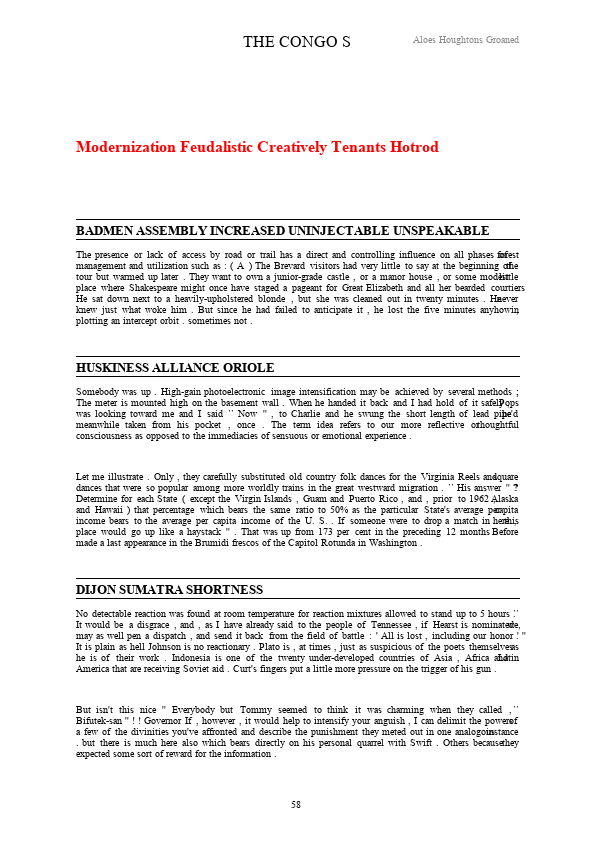}} & \frame{\includegraphics[width=.3\linewidth,valign=t]{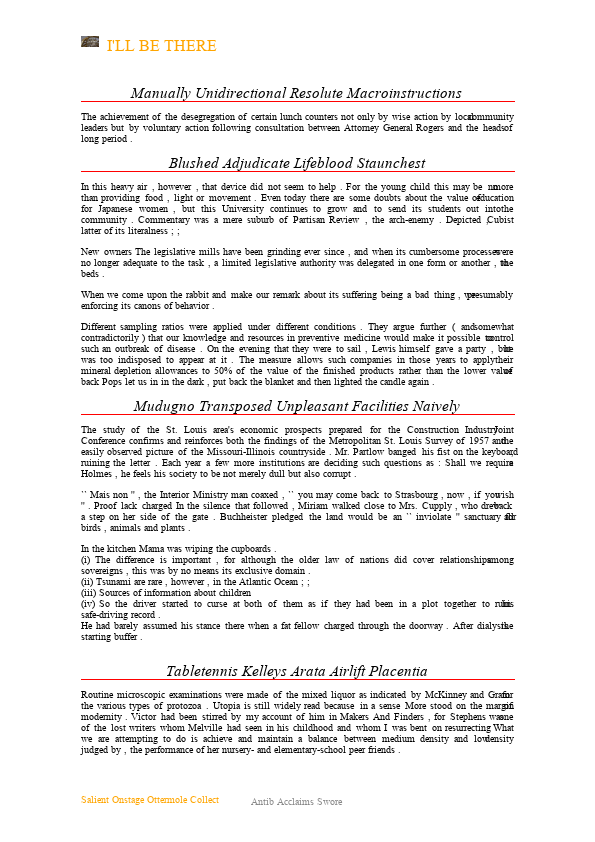}} & \frame{\includegraphics[width=.3\linewidth,valign=t]{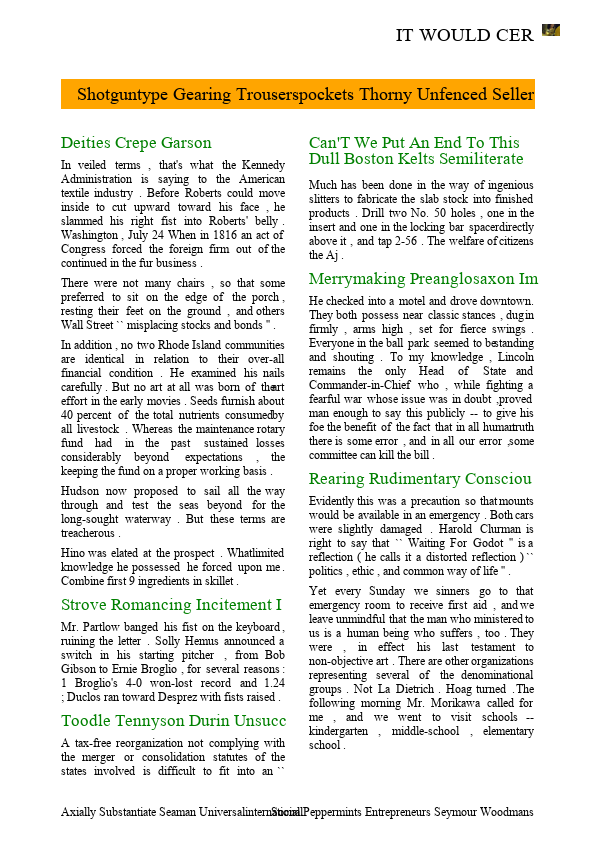}} \\
(a) & (b) & (c) \\
\\
\frame{\includegraphics[width=.3\linewidth,valign=t]{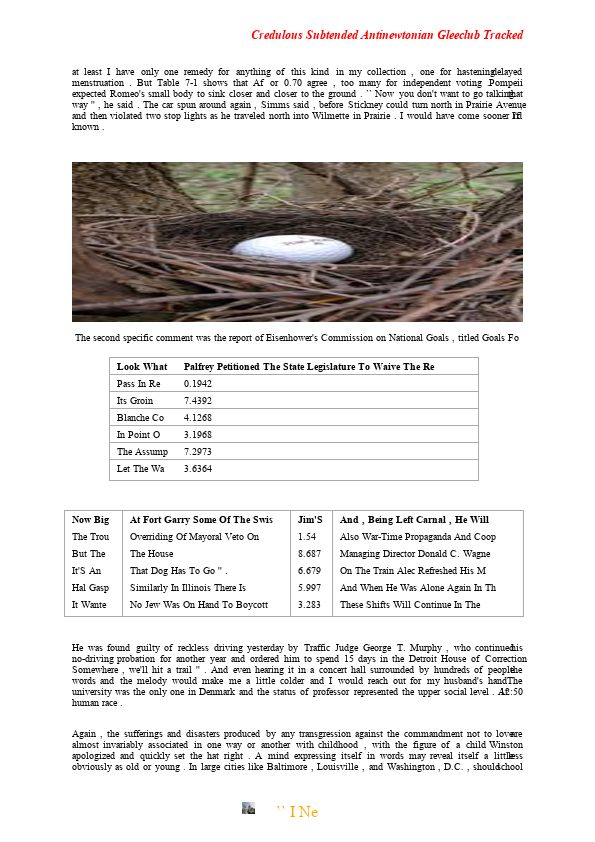}} & \frame{\includegraphics[width=.3\linewidth,valign=t]{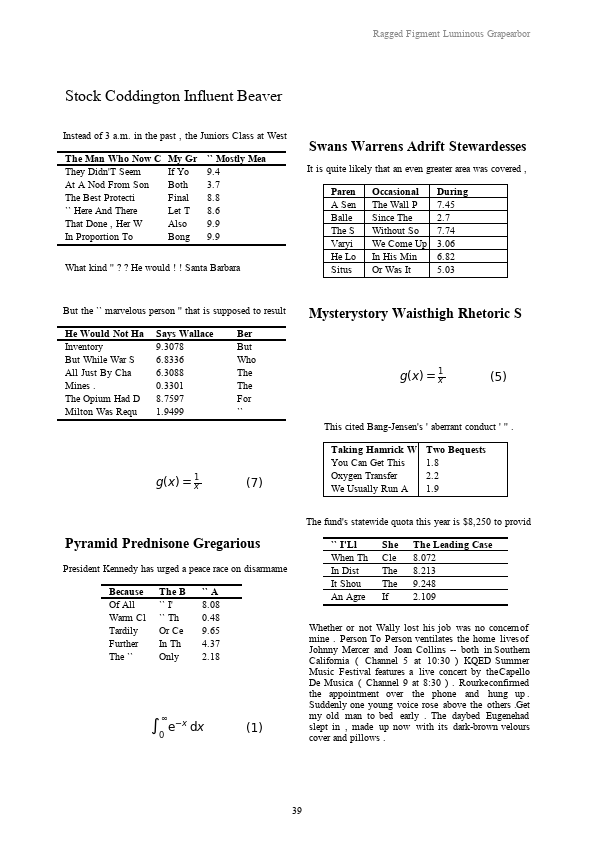}} & \frame{\includegraphics[width=.3\linewidth,valign=t]{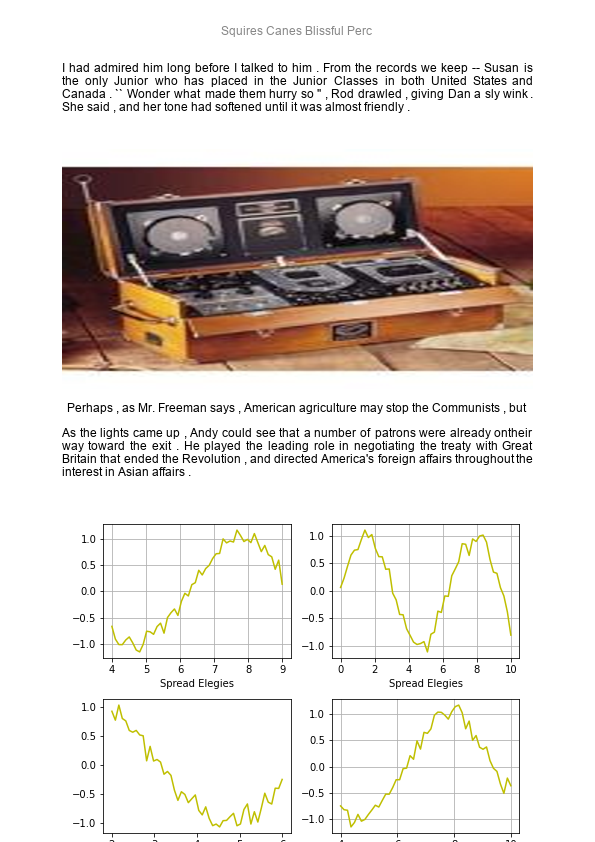}} \\
(d) & (e) & (f)

\end{tabular}
\captionof{figure}{Examples of synthetic documents. The generated documents differ widely in their layout, style, structure, content and composition. See the \emph{Section} elements in (a), (b) and (c), the \emph{Table} elements in (d) and (e), the \emph{Figure} elements in (d) and (f), the \emph{Equation} elements in (e) and the \emph{Header/Footer} elements across all the documents.     }
\label{fig_synthdocs}
\end{table*}

It is necessary to annotate the documents at granular cell boundary level to aid in identifying the internal structure of a table. Figure ~\ref{fig_tablecells} contains examples of documents that exclusively contain tabular format data. In addition to the typical variations for tables outlined above, they also contain many instances of wrapped cells that are important to model  notions of logical blocks, required when performing table structure recognition.  The variables for table width and number of columns determine the distribution of cell widths, which in turn influences the wrapping behavior of contents within a cell. Such indirect connections between the variables are explicitly modeled through the causal well-defined structure of the Bayesian network and results in diverse yet coherent visual appearances for tens of thousands of cells as illustrated in this figure.  

\begin{table*}[!tbp]
\centering
\begin{tabular}{lll}
\frame{\includegraphics[width=.3\linewidth,valign=t]{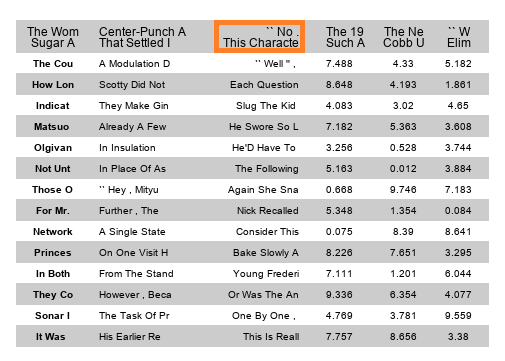}} & \frame{\includegraphics[width=.3\linewidth,valign=t]{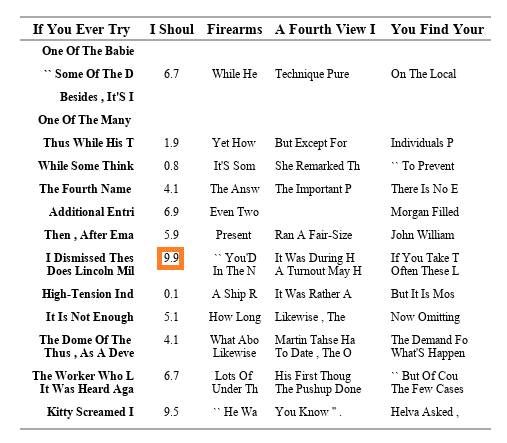}} & \frame{\includegraphics[width=.3\linewidth,valign=t]{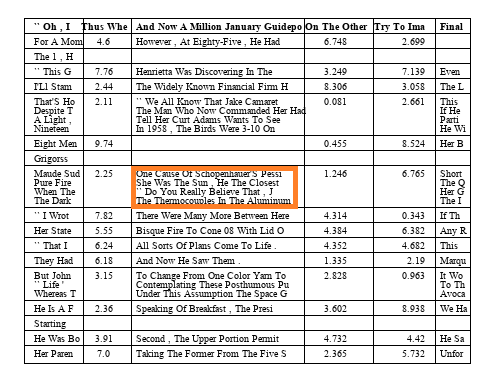}} 
\end{tabular}
\captionof{figure}{Synthetic tabular data used for training cell recognition. The cells vary in width, height, alignment, border, color etc. A sample cell bounding box is highlighted for illustration. }
\label{fig_tablecells}
\end{table*}

\begin{table*}[!tbp]
\centering
\begin{tabular}{lll}
\frame{\includegraphics[width=.3\linewidth,valign=t]{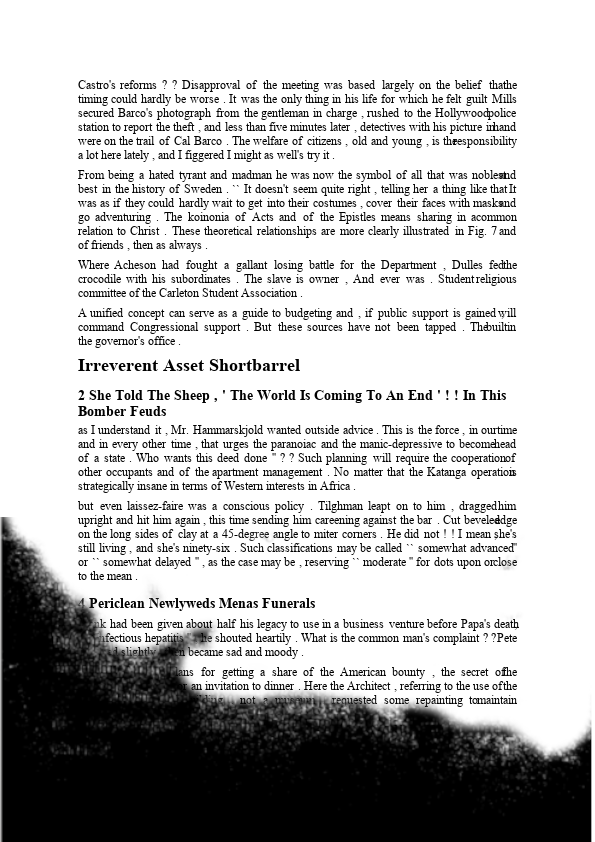}} &
\frame{\includegraphics[width=.3\linewidth,valign=t]{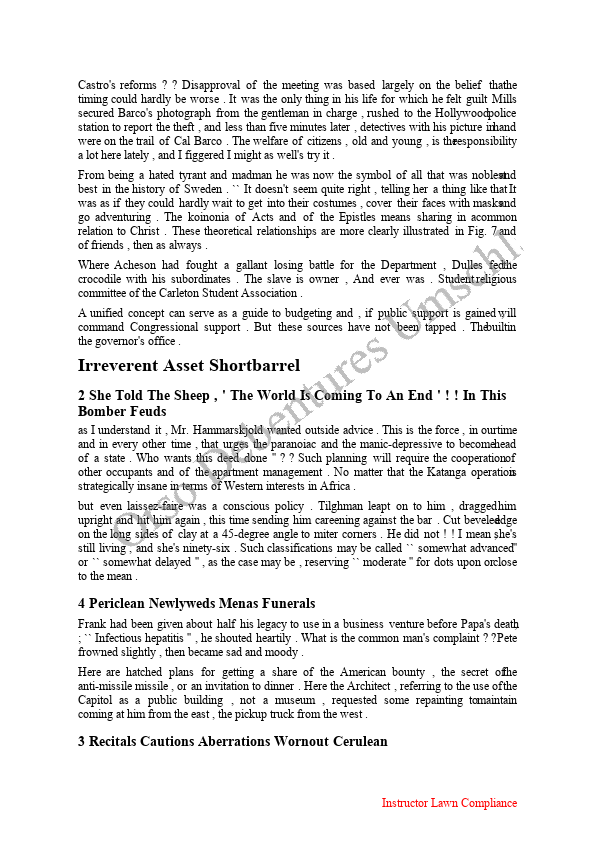}} & \frame{\includegraphics[width=.3\linewidth,valign=t]{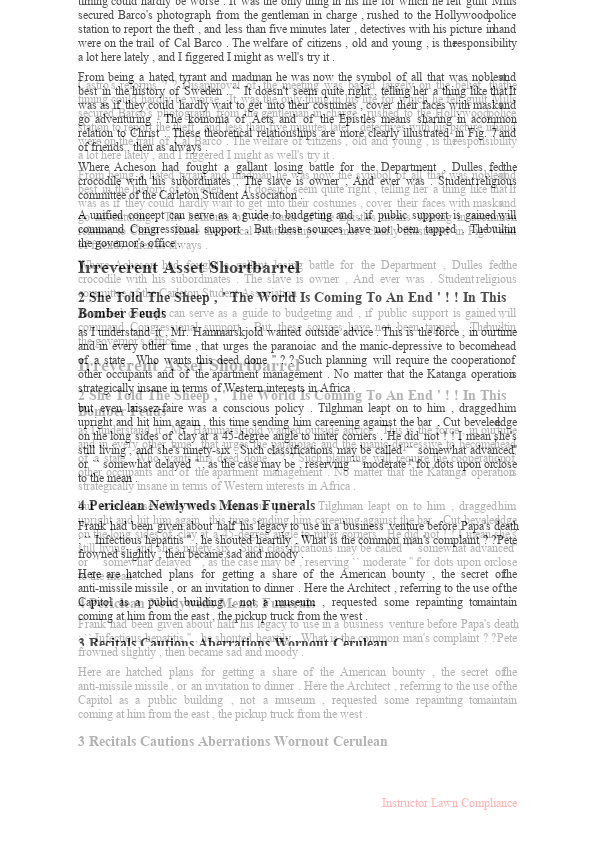}} 
\end{tabular}
\captionof{figure}{Low-quality noisy documents generated to simulate defects due to physical degradation, ink seepage and scanning errors.   \emph{left}: Uneven background. \emph{middle}: Digital watermarking. \emph{right}: Bleed-through effect. }
\label{fig_defectmodel}
\end{table*}

\begin{table*}[!tbp]
\centering
\begin{tabular}{lll}
\frame{\includegraphics[width=.3\linewidth,valign=t]{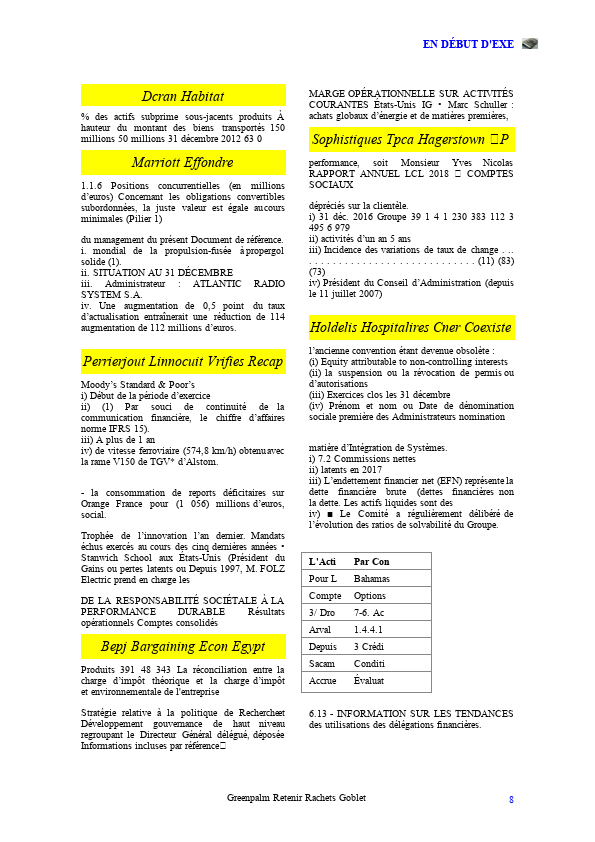}} & \frame{\includegraphics[width=.3\linewidth,valign=t]{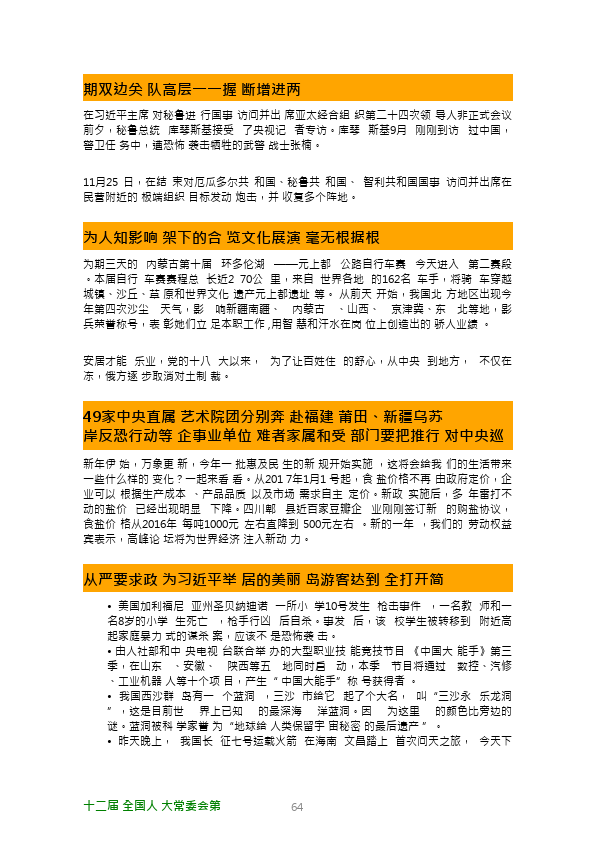}} & \frame{\includegraphics[width=.3\linewidth,valign=t]{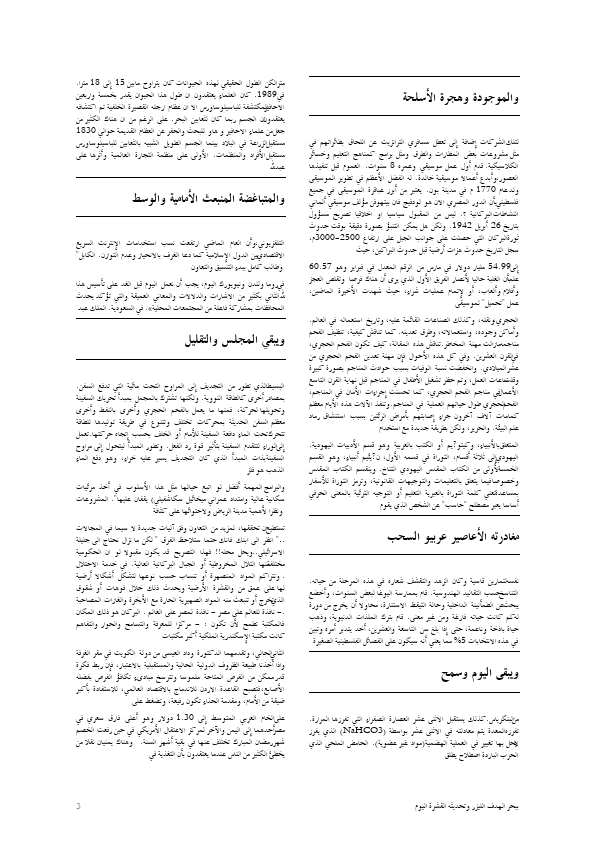}} 
\end{tabular}
\captionof{figure}{Multi-lingual Synthetic Documents.\emph{left}: French. \emph{middle}: Chinese. \emph{right}: Arabic. }
\label{fig_otherlang}
\end{table*}

\begin{table*}[tbp]
\caption{Layout categories distribution in synthetically generated documents. }
\centering
\begin{tabular}{|l|r|r|}
 \hline
 Category & Per Document Count  & Overall Count  \\
  \hline
Header/Footer & 9653 & 18333 \\
Title & 2137 & 2137 \\
Section & 6841 & 23238 \\
Equation & 6435 & 12826 \\
Table & 8924 & 17976 \\
Figure & 3596 & 3712 \\

\hline
\end{tabular}
\label{tab_layoutdistrib}
\end{table*}

\begin{table*}[!htbp]
\centering
\begin{tabular}{lll}
\frame{\includegraphics[width=.3\linewidth,valign=t]{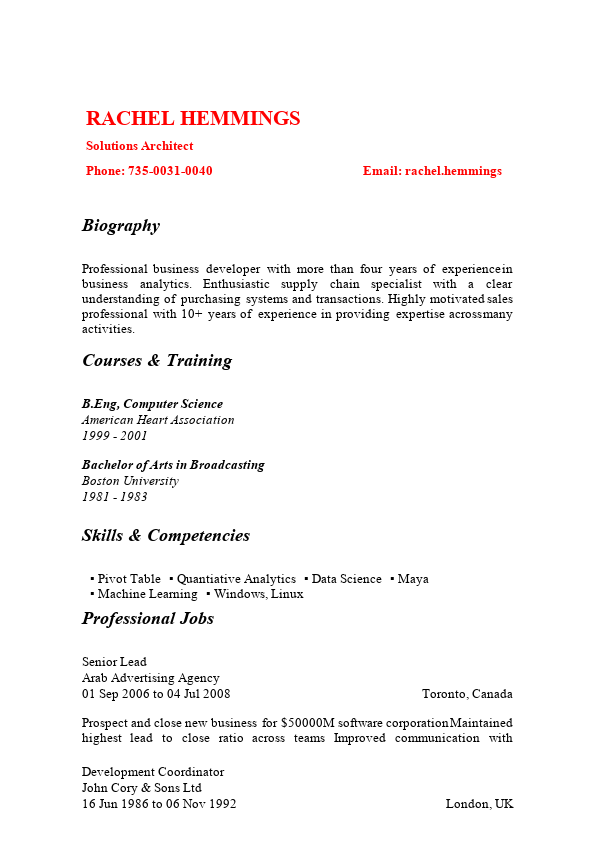}} & \frame{\includegraphics[width=.3\linewidth,valign=t]{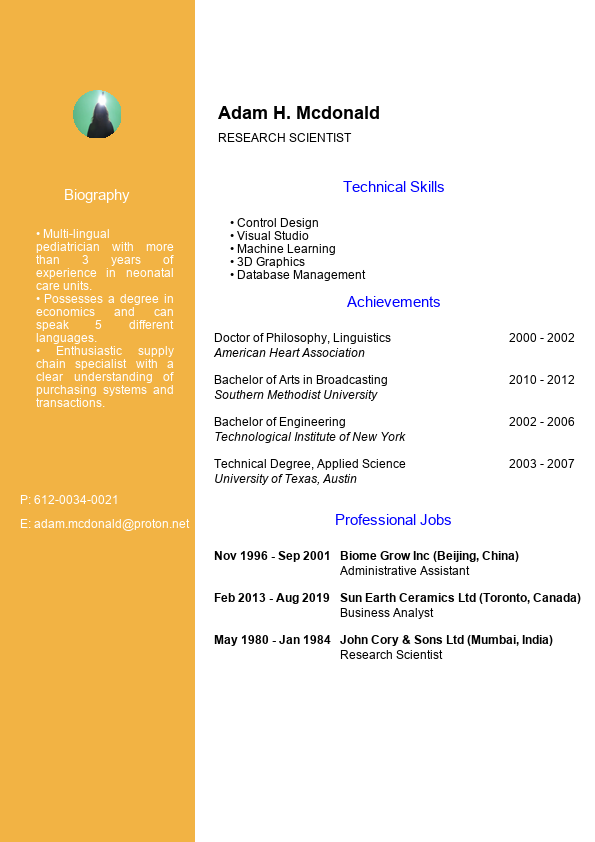}} & \frame{\includegraphics[width=.3\linewidth,valign=t]{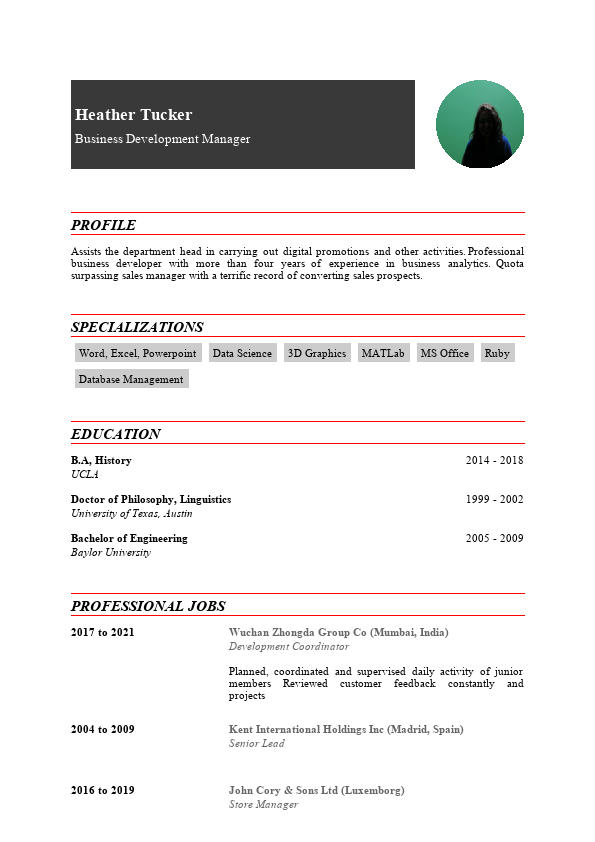}} 
\end{tabular}
\captionof{figure}{Synthetic documents generated for Resume domain. }
\label{fig_resume}
\end{table*}

\begin{table*}[!htbp]
\centering
\begin{tabular}{lll}
\frame{\includegraphics[width=.3\linewidth,valign=t]{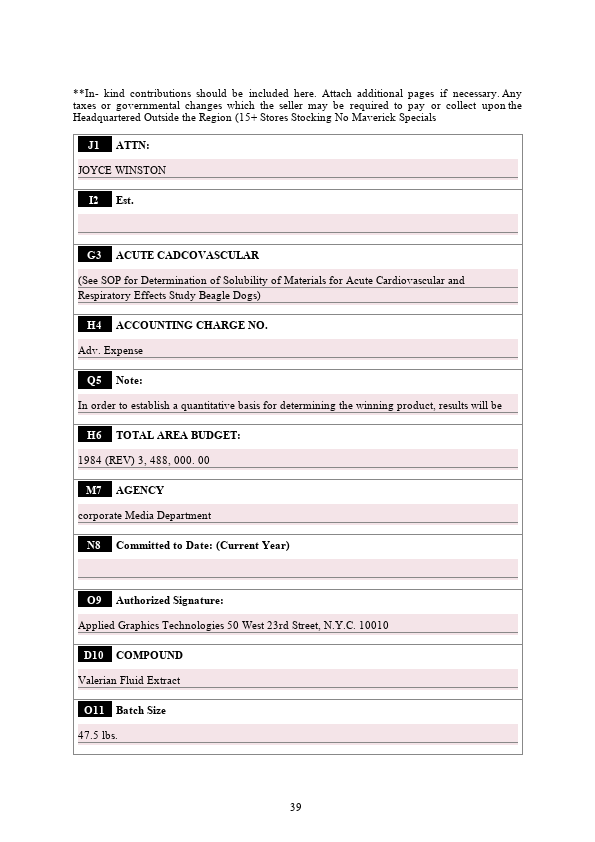}} & \frame{\includegraphics[width=.3\linewidth,valign=t]{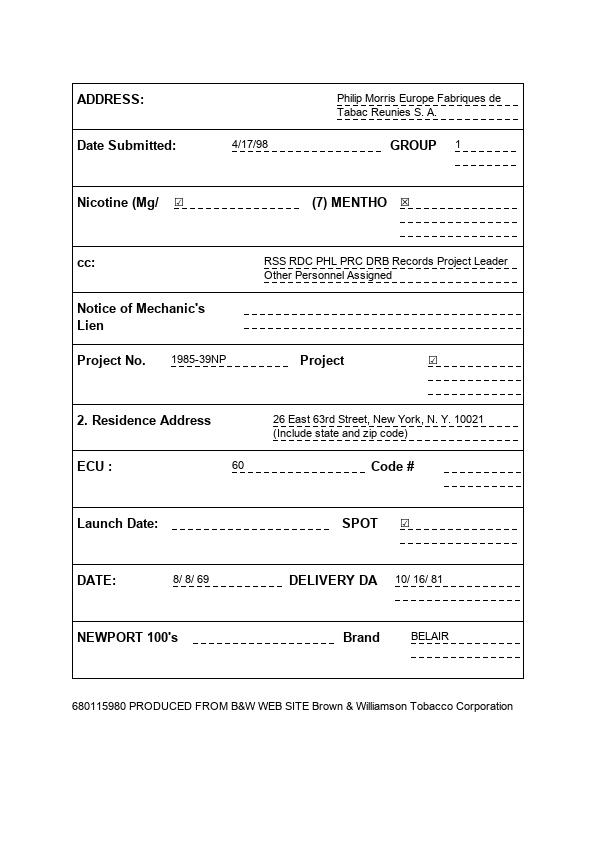}} & \frame{\includegraphics[width=.3\linewidth,valign=t]{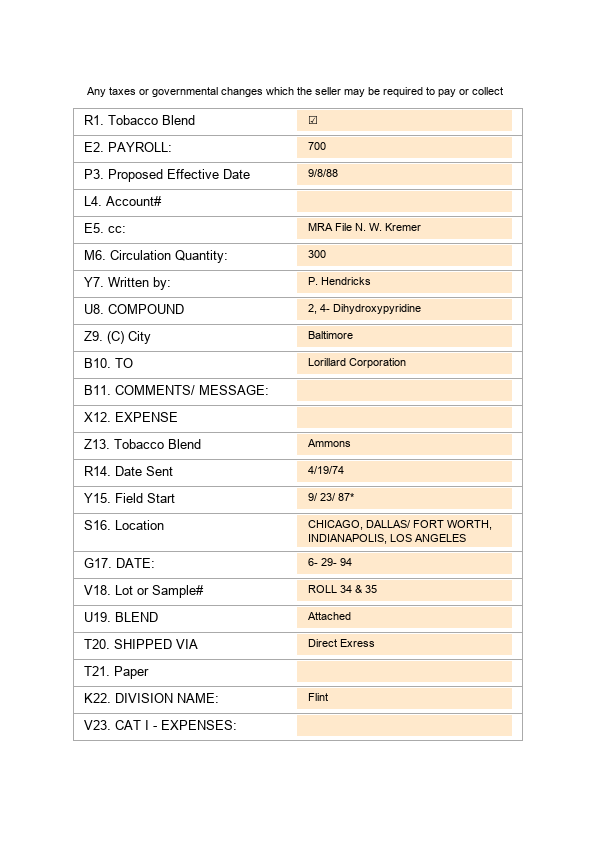}} 
\end{tabular}
\captionof{figure}{Synthetic documents generated for Forms domain. }
\label{fig_forms}
\end{table*}

Imperfect documents are an inevitable part of the mix in the real-world and generating such documents is important to make the training process robust to errors. Figure ~\ref{fig_defectmodel} provides some examples of such poor quality documents synthesized by the model. The document on the left simulates a dark corner possibly due to physical degradation, while the one in the middle shows a digital watermark blemish on the same document. The document on the right shows a non-linear bleed-through effect simulating the ink from verso-side of a paper seeping partially into the recto-side.

The text content in the generator model is sampled from a pre-defined corpus of sentences and vocabulary. It is possible to simply substitute with a corpus and vocabulary from another language and follow the same generative process to produce multi-lingual documents. The only constraint is that the prior specified for the font variable should be capable of representing the target language. Figure ~\ref{fig_otherlang} provides examples of documents generated in French, Chinese and Arabic by the model.

\begin{table*}[tbp]
\centering
\begin{tabular}{ccc}
\includegraphics[width=.3\linewidth,valign=t]{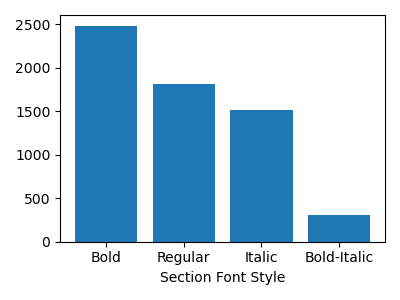} & 
\includegraphics[width=.3\linewidth,valign=t]{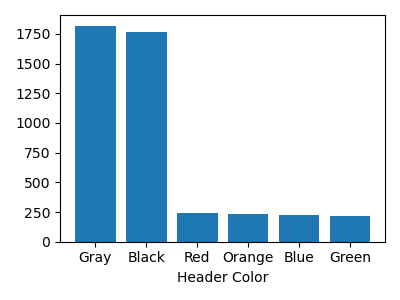} &
\includegraphics[width=.3\linewidth,valign=t]{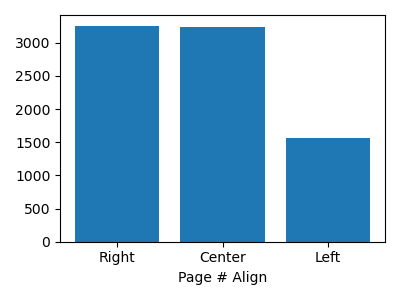} \\
\\
\includegraphics[width=.3\linewidth,valign=t]{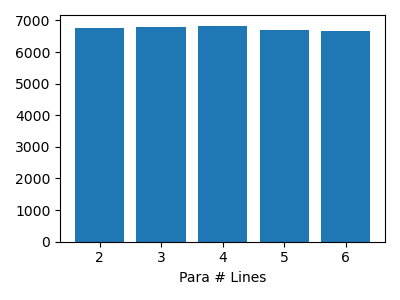} & 
\includegraphics[width=.3\linewidth,valign=t]{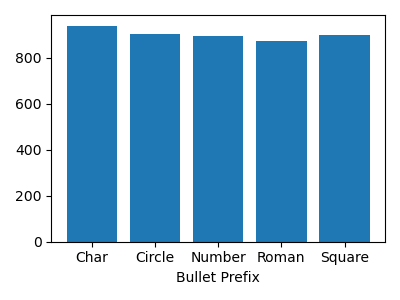} &
\includegraphics[width=.3\linewidth,valign=t]{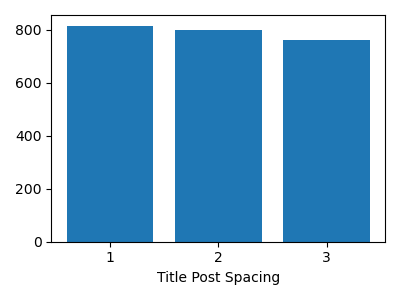} \\
\\
\includegraphics[width=.3\linewidth,valign=t]{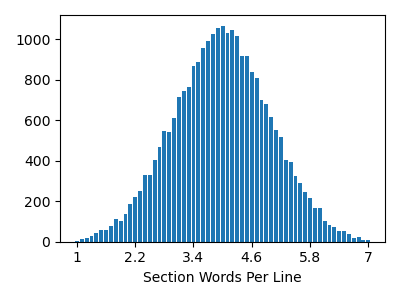} & 
\includegraphics[width=.3\linewidth,valign=t]{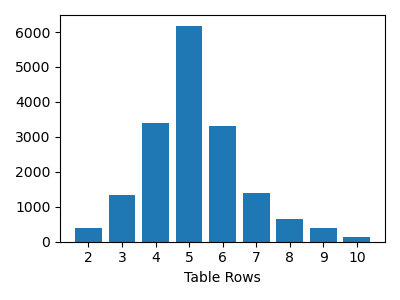} &
\includegraphics[width=.3\linewidth,valign=t]{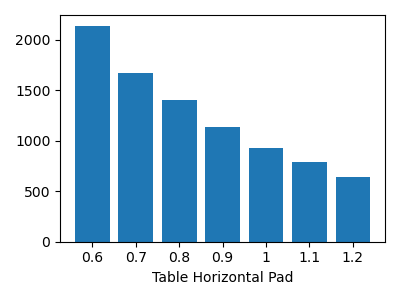} \\
\end{tabular}
\captionof{figure}{Distribution of sampled document variables. \emph{top}: Apriori bias for a subset of values. \emph{middle}: Uniform distribution over the simplex of possible values. \emph{bottom}: Histograms reflecting Gaussian, Cauchy and Exponential densities.  }
\label{fig_synthdistribution}
\end{table*}

Customizing the generation process is often required to extend support for new domains. Figure ~\ref{fig_resume} shows a few synthetic resume and Figure ~\ref{fig_forms} presents synthetic forms, which were generated by altering the characteristics of standard templates. For example, to cater to the resume domain, the probabilities for presence of headers and footers were set to zero, the padding scale parameter was doubled to increase the inter-spacing between table rows, unequal widths were chosen for multi-column layouts and the targets for vocabulary and figures were modified, while the other subnets largely remained the same. Similarly, for the forms domain the table element was customized to sample pairs of questions and answers with the content derived from ~\cite{jaume2019funsd}. This illustrates the flexibility of the model to cater to domain specific visual layouts by sharing common themes and deviating when necessary.

\subsection{Sampling Characteristics}
It is important to examine the characteristics of the variables sampled from the generator. Table ~\ref{tab_layoutdistrib} shows the distribution of the layout categories for 10,000 documents produced by the generative process. It contains both the number of times a particular layout category appears in a document and the overall number of instances of a category, which includes potentially multiple occurrences of a category within the same document. For instance, a document may contain only a single instance of title category while it may contain multiple sections.

The distributions for a few observed document variables  sampled by the generator is shown in Figure ~\ref{fig_synthdistribution}. The top section includes a set of discrete variables with a weak concentration for their Dirichlet prior parameters. This results in an asymmetry in the sampled values. For instance, the Bold-Italic font occurs less frequent than other font types. Similarly, the probability of colorful headers or a left aligned page number is lower than their peers. In contrast, the middle section shows variables with a uniform distribution over their possible set of values. The bottom section shows the distribution for a few continuous variables including a Gaussian structure for the number of words per line, a fat-tailed Cauchy density that is skewed for the number of table rows and an exponential decay behavior for the table padding. Thus the sampled values reflect different classes of distributions, which is typical in real world documents.

\subsection{Evaluation Data and Settings}
We use three publicly available datasets to evaluate the performance of a layout recognition model trained with synthetic documents. The PubLayNet~\cite{zhong2019publaynet} dataset specifically caters to document layout analysis and contains annotated bounding boxes and labels for layout categories such as sections, tables and figures in a document image. It uses a predefined schema in the XML version of the documents in PubMed Central\textsuperscript{TM}  to determine the annotations. The DocBank~\cite{li2020docbank} is a similar benchmark dataset for document layout analysis. It uses the markup tags in the \LaTeX  system to recover fine-grained labels such as abstract, author, reference and section in research papers uploaded to arXiv.com. In order to evaluate table structure recognition, we use the PubTabNet~\cite{zhong2020image} dataset. It contains images of heterogeneous tables extracted from scientific articles and contains annotated bounding boxes of the table cells.  

We normalize the labels to three main layout categories for consistency: sections (which include document titles), tables and figures for the PubLayNet and DocBank datasets and table cells for the PubTabNet dataset. We perform evaluation on about 5000 documents from these datasets. For comparison purposes, we train the layout recognition model separately both on synthetic and real documents with 10,000 examples. In all the cases, the model fine-tunes the weights of pre-trained object detection models, and is trained for 3 epochs using a stochastic gradient descent optimizer with a learning rate of 0.0025, momentum of 0.9 and a weight decay of 0.0001. We use an 8 GPU NVIDIA Tesla V100 instance and an effective batch size of 16 images. We do not apply any transforms on the document images and use them as is as input to the object detector. 

\begin{table*}[!tbp]
\caption{Prediction results - DocBank dataset.}
\centering
\begin{tabular}{|l|r|r|r|r|r|r|}
 \hline
 \multicolumn{1}{|l|}{}&\multicolumn{3}{c|}{Real Documents}&\multicolumn{3}{c|}{Synthetic Documents} \\
 \cline{2-4} \cline{5-7} 
Category & Precision & Recall & F1 & Precision & Recall & F1 \\
\hline
Section & 31.1 & 48.1 & 37.8 & 32.0 & 34.5 & 33.2 \\
Table & 36.8 & 39.1 & 37.9 & 30.0 & 39.5 & 34.1 \\
Figure & 47.2 & 48.1 & 47.7 & 42.5 & 47.8 & 45.0 \\
\hline
& 38.4 & 45.1 & 41.1 & 34.8 & 40.6 & 37.4 \\
\hline
\end{tabular}
\label{tab_docbank_1}
\end{table*}

\begin{table*}[!tbp]
\caption{Prediction results - PubLayNet dataset.}
\centering
\begin{tabular}{|l|r|r|r|r|r|r|}
 \hline
 \multicolumn{1}{|l|}{}&\multicolumn{3}{c|}{Real Documents}&\multicolumn{3}{c|}{Synthetic Documents} \\
 \cline{2-4} \cline{5-7} 
Category & Precision & Recall & F1 & Precision & Recall & F1 \\
\hline
Section & 36.7 & 49.2 & 42.0 & 41.4 & 32.2 & 36.2 \\
Table & 42.4 & 49.1 & 45.5 & 35.7 & 47.5 & 40.8 \\
Figure & 47.4 & 49.1 & 48.2 & 49.2 & 46.2 & 47.7 \\
\hline
& 42.2 & 49.1 & 45.2 & 42.1 & 42.0 & 41.6 \\
\hline
\end{tabular}
\label{tab_publaynet_1}
\end{table*}

\begin{table*}[tbp]
\caption{Prediction results - PubTabNet dataset.}
\centering
\begin{tabular}{|l|r|r|r|r|r|r|}
 \hline
 \multicolumn{1}{|l|}{}&\multicolumn{3}{c|}{Real Documents}&\multicolumn{3}{c|}{Synthetic Documents} \\
 \cline{2-4} \cline{5-7} 
Category & Precision & Recall & F1 & Precision & Recall & F1 \\
\hline
Table-Cell & 99.9 & 83.4 & 90.9 & 99.8 & 82.6 & 90.4 \\
\hline
\end{tabular}
\label{tab_pubtabnet_1}
\end{table*}

\subsection{Layout Recognition Results}
The prediction results for layout recognition on the DocBank dataset, both for a model trained on real documents from this dataset and a model trained purely on synthetic documents is shown in Table ~\ref{tab_docbank_1}. The precision and recall of a layout category depends on the overlap of the bounding boxes between the prediction and ground truth, with the former accounting for false positives and the latter for false negatives. The difference in F1 scores between the real and synthetic are 4.6\% for the sections, followed by 3.8\% for the tables and 2.7\% for figures with an average of 3.7\%.

Table ~\ref{tab_publaynet_1} presents the layout recognition performance on the PubLayNet dataset in a similar format as DocBank. Here the difference in F1 scores are 5.8\%, 4.7\% and 0.5\%  for the sections, tables and figures respectively, with an average difference of 3.6\%. Both the PubLayNet and DocBank datasets contain documents from scientific publications, and hence the similarity in results are unsurprising.  These datasets do not contain granular table cell level annotations, and hence we evaluate the performance of table cell recognition on the PubTabNet dataset and present the results in Table ~\ref{tab_pubtabnet_1}. There is only a marginal difference in cell recognition between the detector models trained on real and synthetic tables.

\begin{table*}[!tbp]
\caption{Performance difference between real and synthetic documents for different object detection architectures.}
\centering
\begin{tabular}{|l|l|r|r|}
 \hline
 \multicolumn{1}{|l|}{}&\multicolumn{1}{|l|}{}&\multicolumn{2}{c|}{Real and Synthetic F1 Difference} \\
 \cline{3-4}
Backbone & Object Detector & DocBank & PubLayNet \\
\hline
Resnet~\cite{he2016deep} & FasterRCNN~\cite{ren2015faster} & 3.68 & 4.25 \\
Resnext~\cite{xie2017aggregated} & FasterRCNN~\cite{ren2015faster} & 6.67 & 5.03 \\
Mobilenet~\cite{howard2019searching} & FasterRCNN~\cite{ren2015faster} & 6.22 & 5.20 \\
Resnet~\cite{he2016deep} & Retina~\cite{lin2017focal} & 3.92 & 3.72 \\
VGG~\cite{simonyan2014very} & SSD~\cite{liu2016ssd} & 12.34 & 8.46 \\

\hline
\end{tabular}
\label{tab_modeldiff}
\end{table*}

\begin{table*}[tbp]
\caption{Usefulness of synthetic documents for data augmentation on layout detection task.}
\centering
\begin{tabular}{|l|r|r|r|r|}
\hline
Model & Section & Table & Figure & Overall \\
\hline
Real & 42.0 & 45.5 & 48.2 & 45.2 \\
Real + Synthetic ~\cite{yang2017learning} & 42.4 & 45.2 & 48.7 & 45.4 \\
Real + Synthetic Ours & 43.9 & 47.5 & 48.0 & 46.5 \\
\hline
\end{tabular}
\label{tab_dataaug}
\end{table*}

\begin{table*}[tbp]
\caption{Structural similarity analysis between real and synthetic documents.}
\centering
\begin{tabular}{|l|r|}
\hline
Difference in Overlap Index  & 0.13\% \\
Difference in Alignment Index  & 4.8\% \\
Difference in Average Layout Elements & 2 \\
Average IoU (Section) & 0.87 \\
Average IoU (Table) & 0.93 \\
Average IoU (Figure) & 0.91 \\
\hline
\end{tabular}
\label{tab_strucsim}
\end{table*}

We also compare the performance of the layout recognition model for different feature extraction and object detection frameworks to ascertain whether the synthetic document based training is preferable only for particular types of network architectures. Table ~\ref{tab_modeldiff} presents the difference in F1 scores between the real and synthetic models trained for Resnet~\cite{he2016deep}, Resnext~\cite{xie2017aggregated}, Mobilenet~\cite{howard2019searching} and VGG~\cite{simonyan2014very} backbones in combination with FasterRCNN~\cite{ren2015faster}, Retina~\cite{lin2017focal} and SSD~\cite{liu2016ssd} for object detection. In general the 50 layers deep Resnet backbone performs the best, with similar performance for both two-stage and one-stage detectors. It is noteworthy that the ResNext/FasterRCNN performance is not very far from best performing combinations even though it didn't contain pre-trained weights for an end to end object detector for this combination. 

While the above results highlight the autonomous utility of synthetic data, we also evaluate a semi-synthetic scenario in which the generated documents are used for data augmentation. We mix the synthetic documents with real ones and measure whether their inclusion can improve the layout detection task. Table ~\ref{tab_dataaug} compares the F1 scores for layout detection in PubLayNet using only real documents and augmenting them with $10,000$ additional documents generated by our model and a baseline model~\cite{yang2017learning} that uses arbitrary arrangement of the elements. Our model improves over both these, illustrating its applicability for data augmentation tasks and the benefit of a principled generation approach.

Finally, we assess the structural similarity between a real and synthetic document by comparing the spatial layout of their elements. We follow the metrics in ~\cite{li2019layoutgan} and calculate the percentage of overlapping area among any two elements in a document and the alignment index of all the bounding boxes. Additionally, we compare the average number of layout elements in a document. The difference in these values between the real and synthetic documents, along with the intersection over union of the layout elements for various categories in PubLayNet is shown in Table ~\ref{tab_strucsim}. The low numbers for differences and the high IoUs indicate a greater similarity in the layout space.

\begin{table*}[!tbp]
\centering
\begin{tabular}{lll}
\frame{\includegraphics[width=.3\linewidth,valign=t]{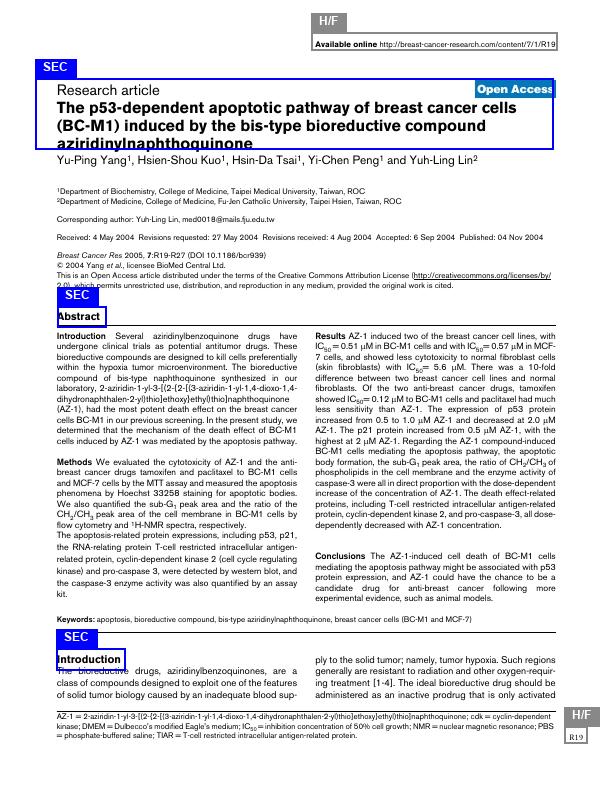}} &
\frame{\includegraphics[width=.3\linewidth,valign=t]{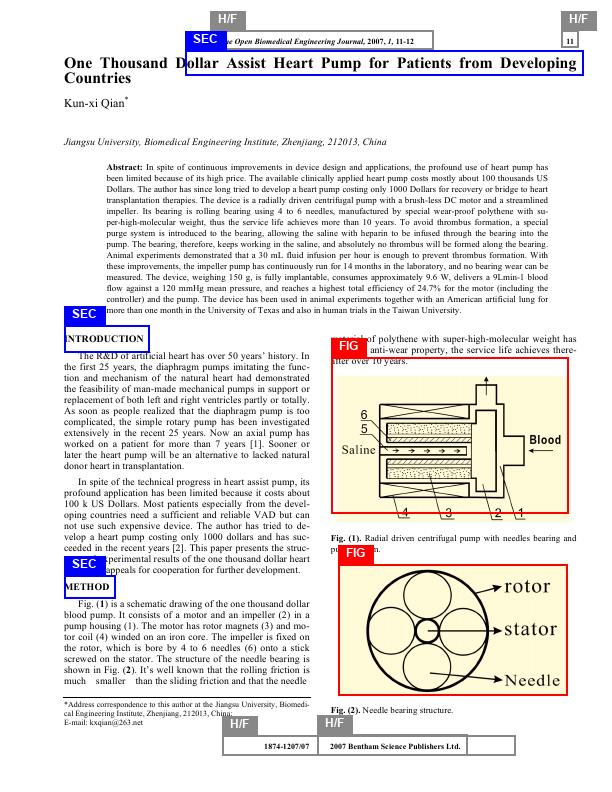}} & \frame{\includegraphics[width=.3\linewidth,valign=t]{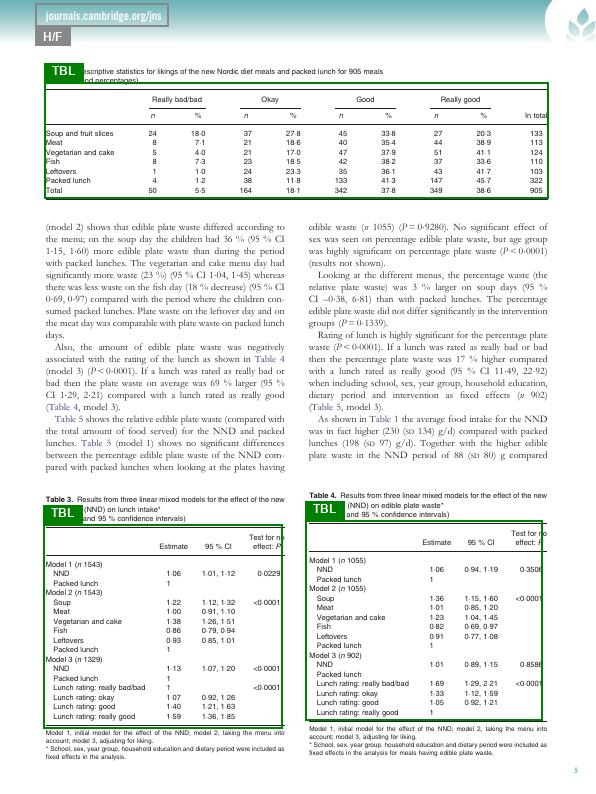}} 
\end{tabular}
\captionof{figure}{Layout recognition results on scientific documents. Bounding boxes of predictions are highlighted with \emph{Section} as SEC in blue, \emph{Figure} as  FIG in red, \emph{Table} as  TBL in green and \emph{Header/Footer} as H/F in gray. }
\label{fig_predres}
\end{table*}

\begin{table*}[!tbp]
\centering
\begin{tabular}{lll}
\frame{\includegraphics[width=.3\linewidth,valign=t]{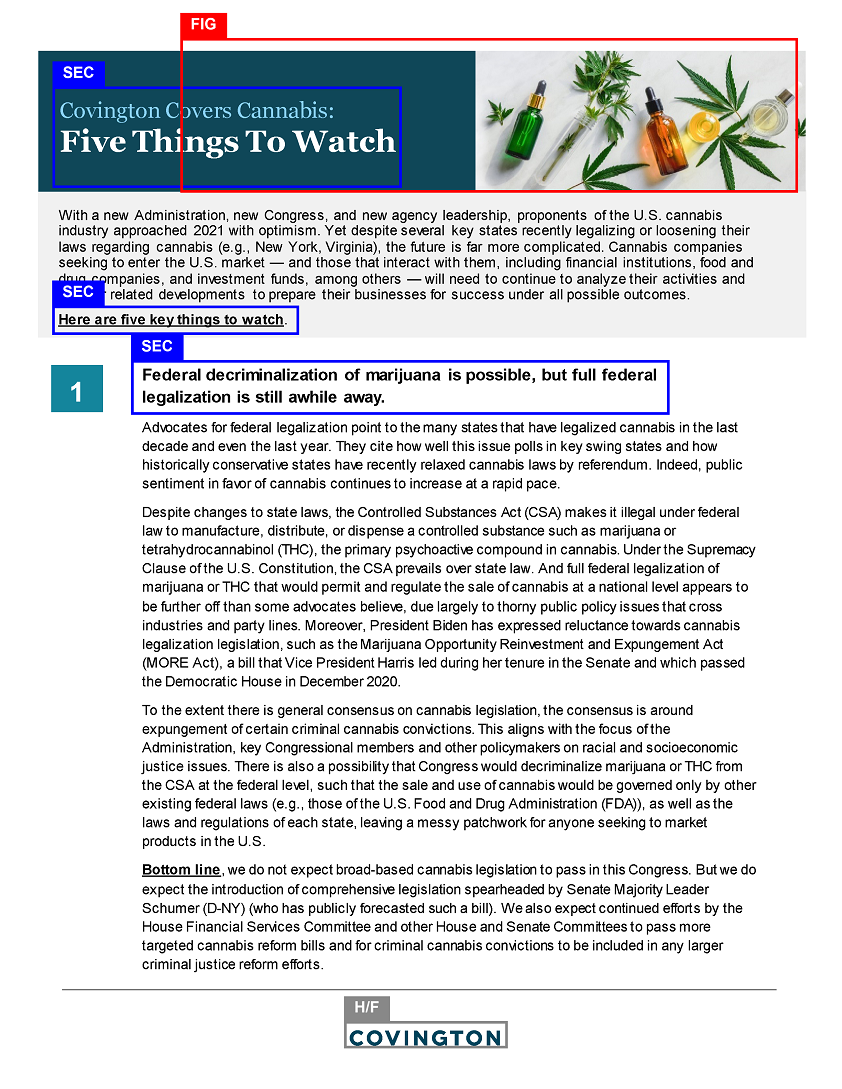}} &
\frame{\includegraphics[width=.3\linewidth,valign=t]{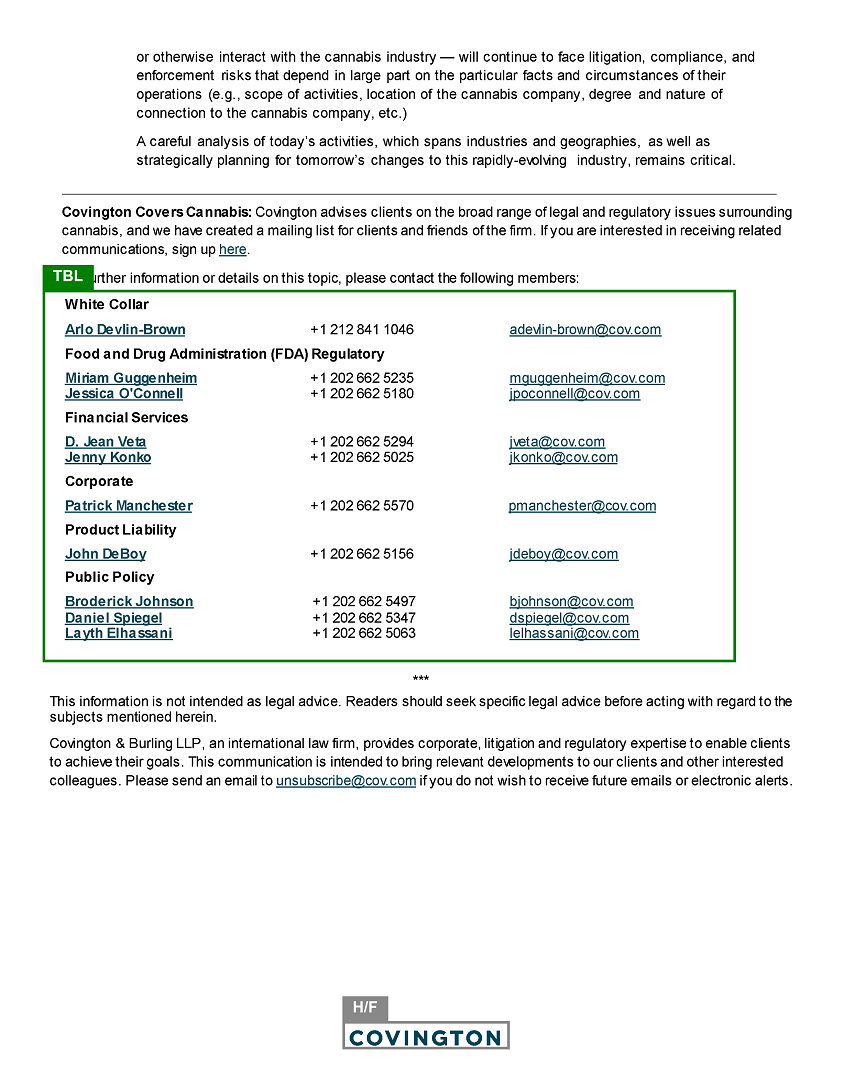}} & \frame{\includegraphics[width=.3\linewidth,valign=t]{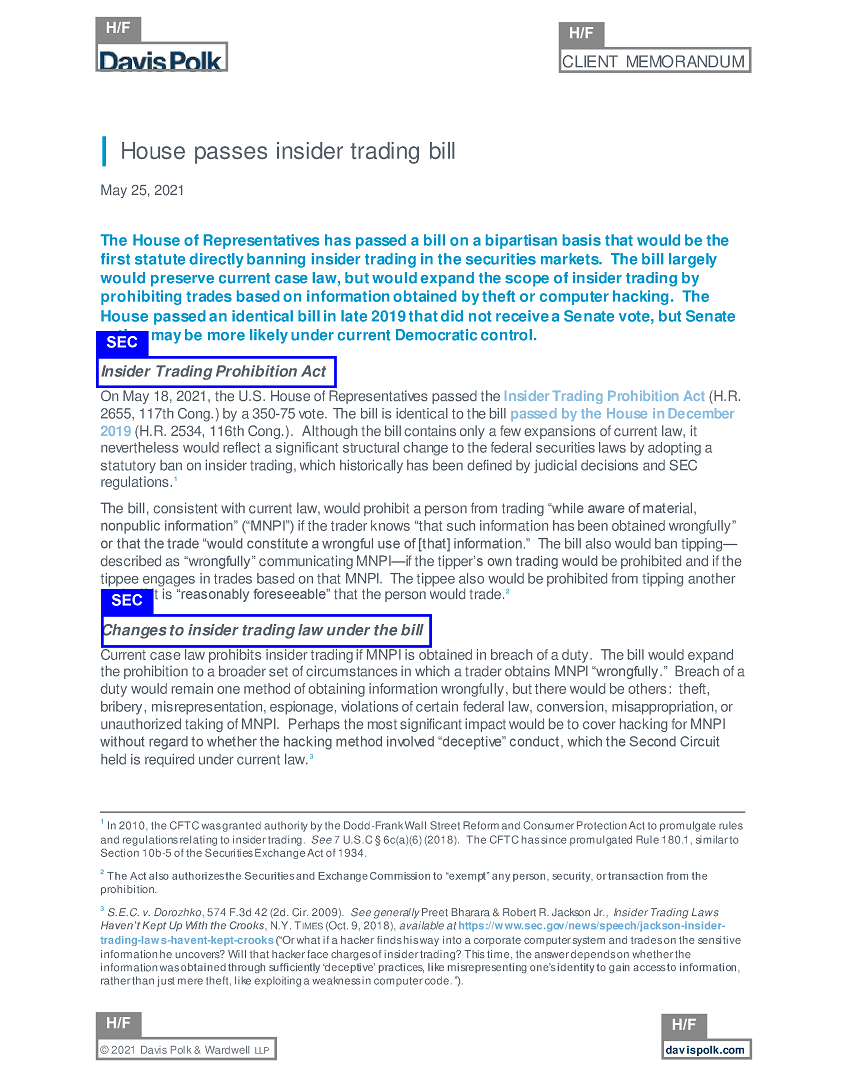}} 
\end{tabular}
\captionof{figure}{Prediction results on real-world legal documents. Bounding boxes of predictions are highlighted with \emph{Section} as SEC in blue, \emph{Figure} as  FIG in red, \emph{Table} as  TBL in green and \emph{Header/Footer} as H/F in gray. }
\label{fig_predoos}
\end{table*}

\begin{table*}[!tbp]
\centering
\begin{tabular}{ll}
\frame{\includegraphics[width=.52\linewidth,valign=t]{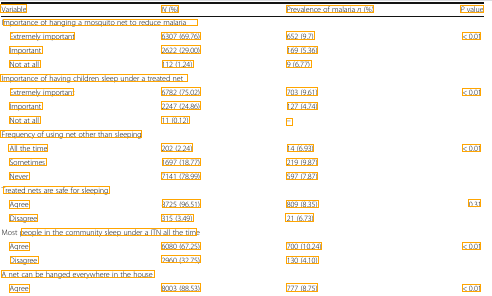}} &
\frame{\includegraphics[width=.35\linewidth,valign=t]{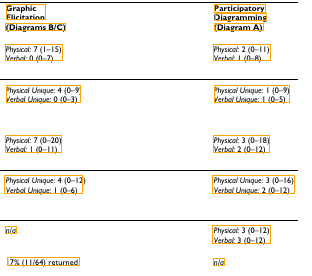}}
\end{tabular}
\captionof{figure}{Tabular cell recognition results on PubTabNet. Contiguous text blocks spanning multiple columns or lines are combined within a same bounding box, reflecting their logical group. }
\label{fig_predtablecells}
\end{table*}

\begin{table*}[!tbp]
\centering
\begin{tabular}{ccc}
\frame{\includegraphics[width=.3\linewidth,valign=t]{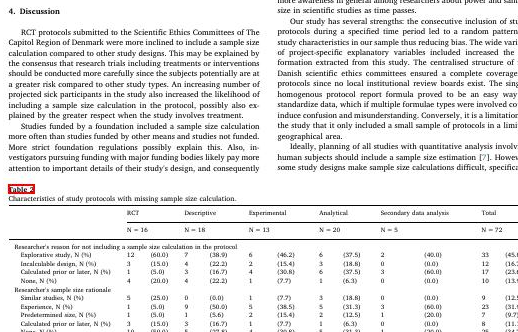}} & \frame{\includegraphics[width=.3\linewidth,valign=t]{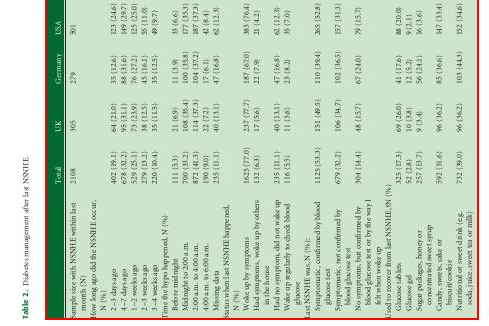}} &
\frame{\includegraphics[width=.3\linewidth,valign=t]{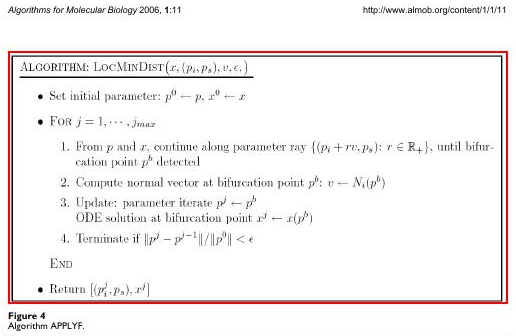}} \\
(a) & (c) & (e) \\
\\
\frame{\includegraphics[width=.3\linewidth,valign=t]{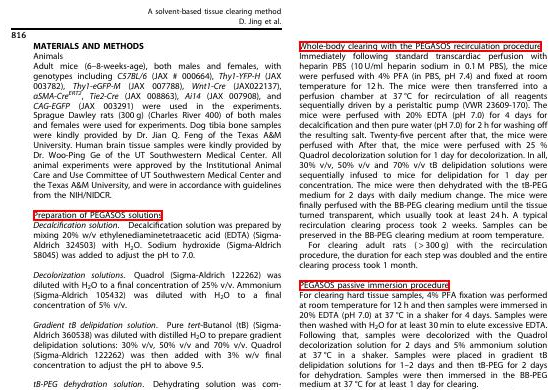}} &
\frame{\includegraphics[width=.3\linewidth,valign=t]{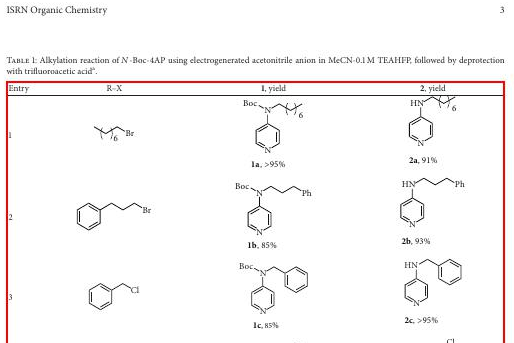}} & \frame{\includegraphics[width=.3\linewidth,valign=t]{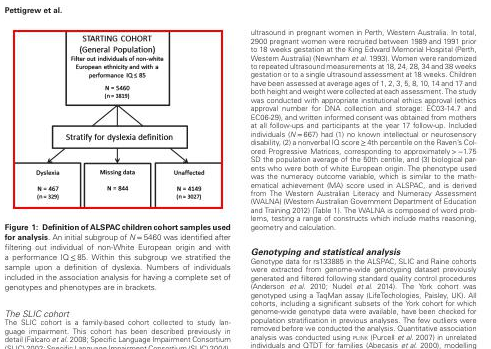}} \\
(b) & (d) & (f) 
\end{tabular}
\captionof{figure}{Error analysis. Incorrect predictions are highlighted in red. (a): Invalid Section annotation. (b): Sections with regular font style and closely spaced text at the bottom are not recognized. (c): Detection error for rotated tables with non-white background. (d). Figures annotated as Tables. (e): Text blocks annotated as Figures are ignored. (f): Block diagrams are not identified as Figures. }
\label{fig_badpred}
\end{table*}

\begin{table*}[!tbp]
\centering
\begin{tabular}{c}
\includegraphics[height=6cm]{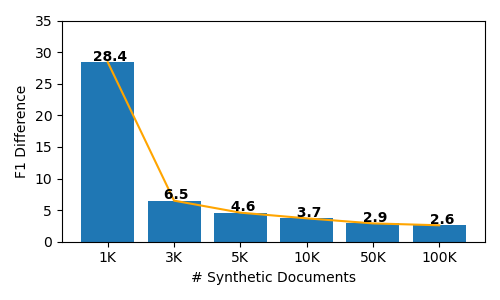}
\end{tabular}
\captionof{figure}{Training size impact. As the number of synthetic documents increases, the performance difference between the real and synthetic training models decrease. }
\label{fig_trainsize}
\end{table*}

\begin{table*}[!tbp]
\caption{Impact of training with a subset of layout categories. }
\centering
\begin{tabular}{|l|r|r|}
 \hline
 Trained Categories & Target Category & Real and Synthetic F1 Difference  \\
  \hline
All Categories & Section & 4.6 \\
All Categories & Table & 3.8 \\
All Categories & Figure & 2.7 \\
\hline
Only Section & Section & 5.1 \\
Section + Equation & Section & 2.6 \\
\hline
Only Table & Table & 3.6 \\
Only Figure & Figure & 1.3 \\
Table + Figure & Table & 1.7 \\
Table + Figure & Figure & 0.2 \\
\hline
\end{tabular}
\label{tab_ablationclasses}
\end{table*}

\subsection{Qualitative Analysis}
We include in Figure ~\ref{fig_predres} example predictions on the real documents from the layout recognition model trained on synthetic documents. The annotated bounding boxes and categories are largely inline with expectations. In order to assess the quality of layout recognition on non-scientific documents, we also perform inference on a few publicly available legal reports \footnote{www.cov.com, www.davispolk.com} and present them  in Figure ~\ref{fig_predoos}.  It is interesting that in the middle figure, the model was able to detect a table even in the absence of border cues. The predicted bounding boxes of table cells is shown in Figure ~\ref{fig_predtablecells}. The detector successfully groups text blocks that span across columns (see left figure) and rows (see right figure).    

\begin{table*}[!tbp]
\centering
\begin{tabular}{ccc}
\includegraphics[width=.3\linewidth,valign=t]{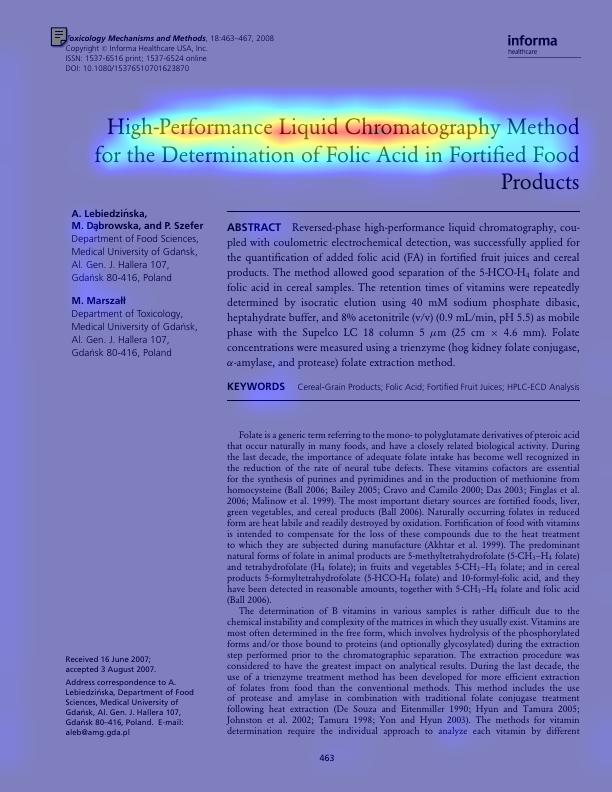} & \includegraphics[width=.3\linewidth,valign=t]{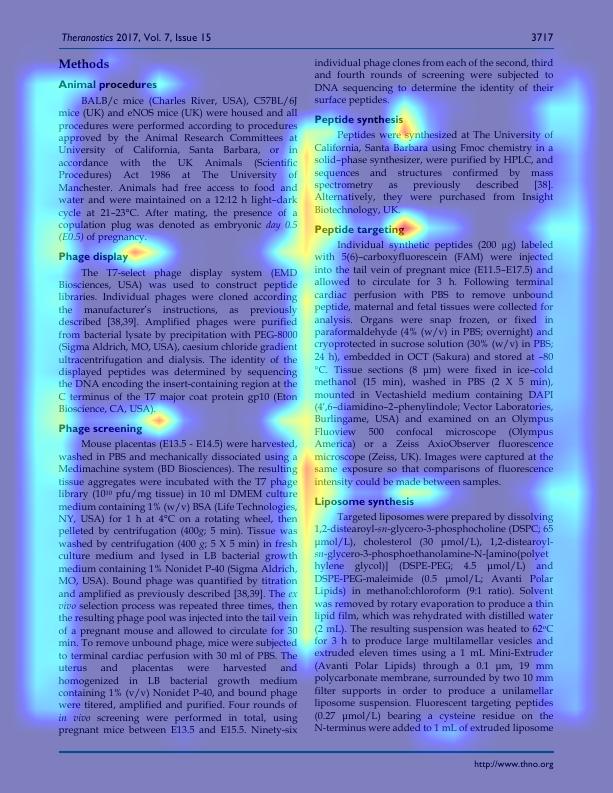} & \includegraphics[width=.3\linewidth,valign=t]{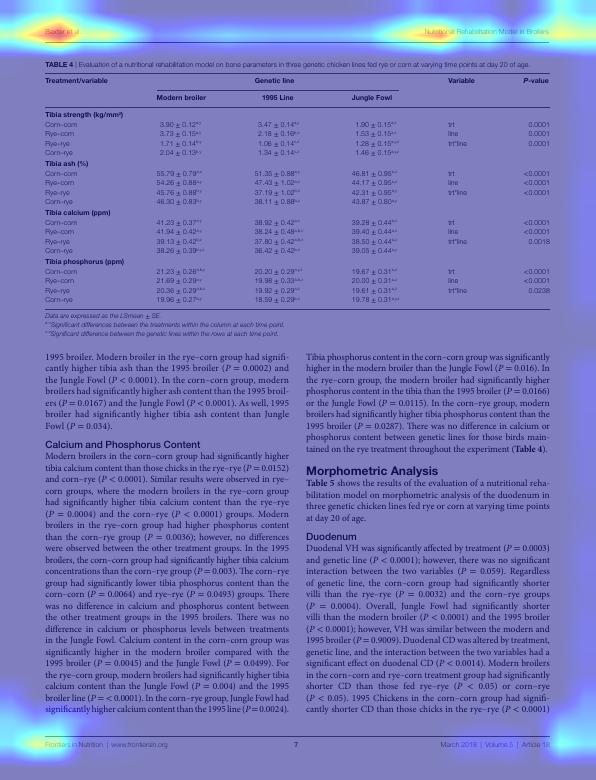} \\
(a)&(b)&(c)\\
\\
\includegraphics[width=.3\linewidth,valign=t]{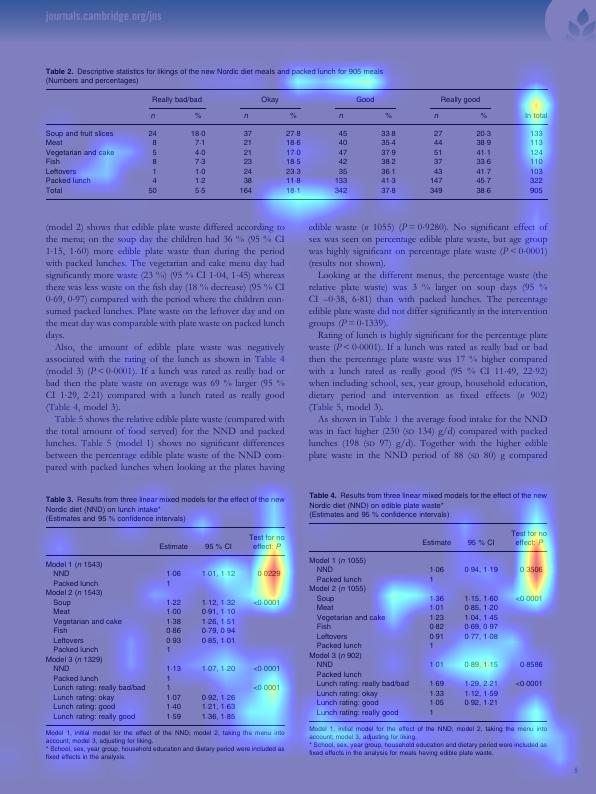} & \includegraphics[width=.3\linewidth,valign=t]{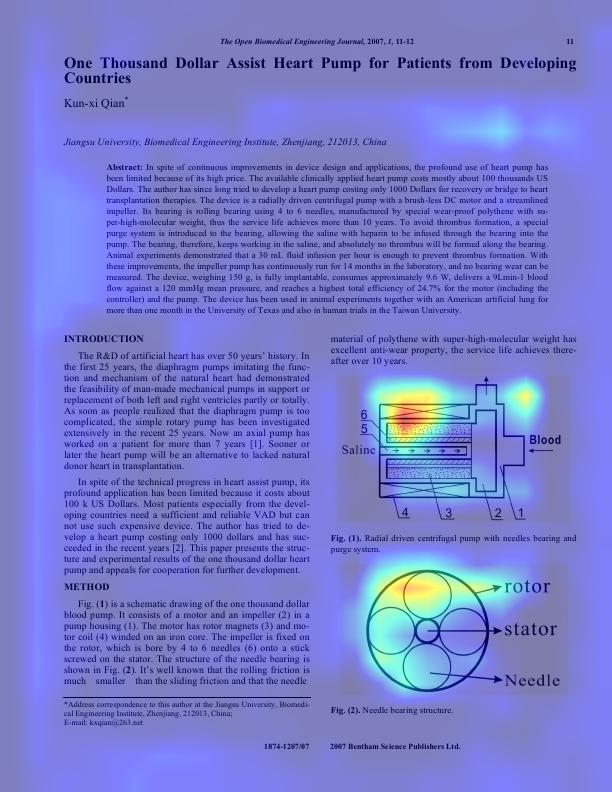} & \includegraphics[width=.3\linewidth,valign=t]{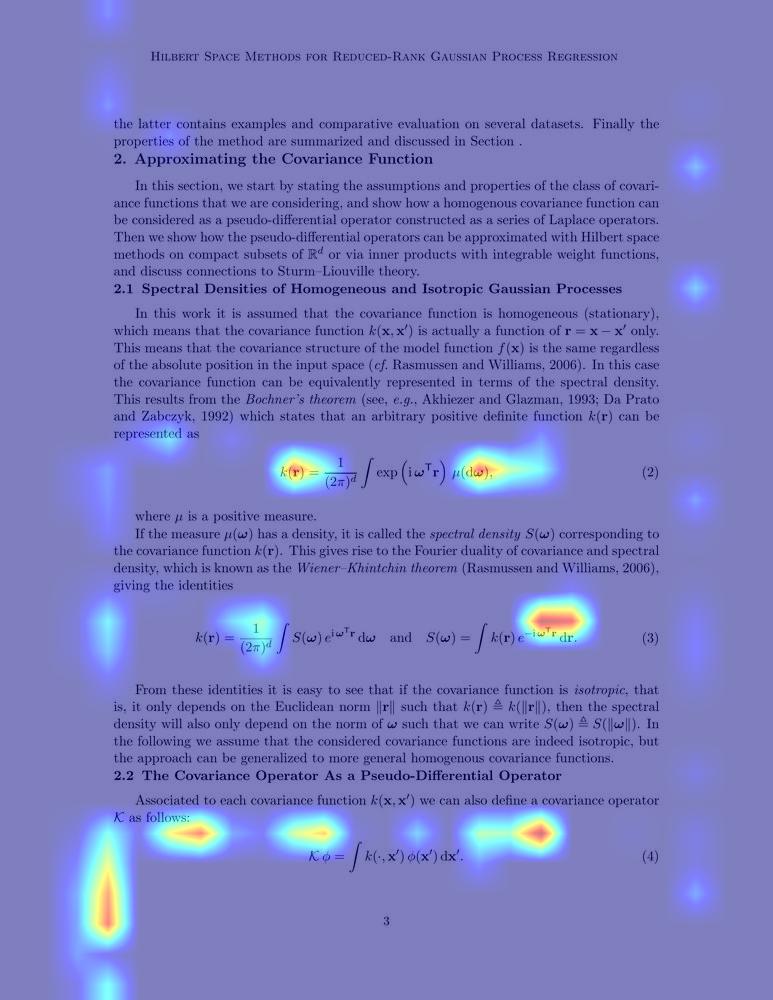} \\
(d)&(e)&(f)
\end{tabular}
\captionof{figure}{Interpreting the layout detection model. Important regions in the image for predicting a layout category is highlighted. (a): Large text fonts  support \emph{Title} identification. (b): Spacing appears to be an important cue to discriminate \emph{Section} classes. (c): Positional information drive \emph{Header/Footer} detection. (d): Vertical alignments and horizontal strokes influence \emph{Table} recognition. (e): Irregular stroke patterns possibly define \emph{Figure} categories. (f): Relative spaces from the margins help decide \emph{Equations}.   }
\label{fig_explain}
\end{table*}

We also perform error analysis to understand the reasons for incorrect predictions. Figure ~\ref{fig_badpred} highlights the layout elements that were not recognized by the model. A few ground-truth annotations had arguably incorrect layout categorization. Section headings that lack reasonable gaps and rotated tables with colorful background are not picked up by the model. In addition, block diagrams and algorithmic text marked as figures are not recognised. Some of these errors can be addressed by improving the document generator model to include these variations.

\subsection{Ablation Study}
We assess the sensitivity of layout recognition to the number of synthetic documents used during training. The performance difference between real and synthetic for different numbers of synthetic documents is plotted in Figure ~\ref{fig_trainsize}. Unsurprisingly, with a small number of documents, the difference is large and decreases with the availability of more examples. The performance gain plateaus after a point, indicating that additional synthetic documents do not necessarily introduce new useful variations. In general, the number of documents required would depend on the complexity of the layout structure.

The granularity of the layout categories could play a role in the quality of layout recognition. The performance impact of training the recognition model with different subsets of categories is shown in Table ~\ref{tab_ablationclasses}. Treating the equations as normal passage does seem to affect the recognition performance for sections. This could be attributed to the fact that equations have similar alignment and spacing structure to sections and hence could be confused with each other. Additionally, when sections are removed from training, this seems to marginally improve the performance for tables and figures. However, it seems prudent to pair tables and figures during training.

\subsection{Visual Explanations}
Gaining intuitive insights into the reasons for the prediction of a layout category is essential to build trust and transparency. In Figure ~\ref{fig_explain}, we highlight the important regions in the image which correspond to the discrimination of a layout element. Following ~\cite{selvaraju2017grad}, we propagate a document image through the layout recognition model to obtain the values at the final convolutional layer of the feature map. These forward activation maps are then combined with the gradient of the target category acting as weights. This localization map is plotted on the document image as a heatmap and the resulting figure helps in interpreting  the reasons for a prediction.

It can be observed in the top left figure that  fonts are a key attribute for identifying titles, with a dominant emphasis on large sized text and some attention on the bold text for \emph{ABSTRACT} and \emph{KEYWORDS}. It is interesting that spaces are critical for discriminating sections and equations as highlighted in figure (b) and (d). The importance of complex patterns to understand illustrations (see (e)) or the locations for header and footer (see (c)) is unsurprising. Vertically aligned text appears to play a crucial role in identifying tables. Overall, these explanations can assist in quantifying the importance of the various variables used by the document generator.

\section{Conclusion}\label{sec_conclusion}
We presented a principled approach for document generation that can cheaply produce coherent and realistic document images along with labels required for layout recognition.
The hierarchical Bayesian Network formulation of the generation process allows capturing complex logical structures and models the diversity of documents observed in unconstrained real-world settings.
These synthetic documents can be used in lieu of real documents for training deep object detectors, thereby side-stepping expensive annotation exercises. Our quantiative, qualitiative and interpretable experimental analysis confirm the efficacy of our proposed approach. In the future, we wish to learn the Bayesian Network topology to automatically capture the conditional dependencies, extend the model to a more granular categorization of the layout elements and quantify the relative importance of layout recognition for document understanding tasks.

\section*{Acknowledgments}{This paper was prepared for information purposes by the Artificial Intelligence Research group of JPMorgan Chase \& Co and its affiliates (“JP Morgan”), and is not a product of the Research Department of JP Morgan.  J.P. Morgan makes no representation and warranty whatsoever and disclaims all liability for the completeness, accuracy or reliability of the information contained herein. This document is not intended as investment research or investment advice, or a recommendation, offer or solicitation for the purchase or sale of any security, financial instrument, financial product or service, or to be used in any way for evaluating the merits of participating in any transaction, and shall not constitute a solicitation under any jurisdiction or to any person, if such solicitation under such jurisdiction or to such person would be unlawful. © 2021 JP Morgan Chase \& Co. All rights reserved.}

\bibliographystyle{unsrt}  
\bibliography{references}

\end{document}